\documentclass[acmsmall]{acmart}
\AtBeginDocument{%
  }

\setcopyright{acmlicensed}
\acmJournal{PACMMOD}
\acmYear{2025} \acmVolume{3} \acmNumber{3 (SIGMOD)}
\acmArticle{201} \acmMonth{6}
\acmDOI{10.1145/3725338}
\received{October 2024}
\received[revised]{January 2025}
\received[accepted]{February 2025}

\usepackage{xspace}
\usepackage{subfigure}
\usepackage{makecell}
\usepackage{multirow}
\usepackage{enumitem}
\usepackage[ruled,linesnumbered]{algorithm2e}
\usepackage{pifont}
\SetKwInOut{Input}{Input}
\SetKwInOut{Output}{Output}
\SetKwInOut{State}{State}
\let\oldnl\nl
\newcommand{\nonl}{\renewcommand{\nl}{\let\nl\oldnl}}

\newcommand{\name}{PQCache\xspace}

\begin{document}

\title{\name: Product Quantization-based KVCache for Long Context LLM Inference}

\author{Hailin Zhang}
\authornote{Hailin Zhang, Xiaodong Ji, Yilin Chen, Fangcheng Fu, Xiaonan Nie, and Bin Cui are with the School of Computer Science \& Key Lab of High Confidence Software Technologies (MOE), Peking University. Bin Cui is also with the Institute of Computational Social Science, Peking University (Qingdao).}
\email{z.hl@pku.edu.com}
\affiliation{%
  \institution{Peking University}
  \country{China}
}

\author{Xiaodong Ji}
\authornotemark[1]
\email{xiaodong.0731@stu.pku.edu.cn}
\affiliation{%
  \institution{Peking University}
  \country{China}
}

\author{Yilin Chen}
\email{yilinology@gmail.com}
\authornotemark[1]
\affiliation{%
  \institution{Peking University}
  \country{China}
}

\author{Fangcheng Fu}
\email{ccchengff@pku.edu.cn}
\authornotemark[1]
\affiliation{%
  \institution{Peking University}
  \country{China}
}

\author{Xupeng Miao}
\email{xupeng@purdue.edu}
\affiliation{%
  \institution{Purdue University}
  \country{USA}
}

\author{Xiaonan Nie}
\email{xiaonan.nie@pku.edu.cn}
\authornotemark[1]
\affiliation{%
  \institution{Peking University}
  \country{China}
}

\author{Weipeng Chen}
\email{hitcsgavin@gmail.com}
\affiliation{%
  \institution{Baichuan Inc.}
  \country{China}
}

\author{Bin Cui}
\email{bin.cui@pku.edu.cn}
\authornotemark[1]
\authornote{Bin Cui is the corresponding author.}
\affiliation{%
  \institution{Peking University}
  \country{China}
}

\renewcommand{\shortauthors}{Hailin Zhang et al.}


\begin{abstract}

As the field of Large Language Models (LLMs) continues to evolve, the context length in inference is steadily growing.
Key-Value Cache (KVCache), the intermediate representations of tokens within LLM inference, has now become the primary memory bottleneck due to limited GPU memory. 
Current methods selectively determine suitable keys and values for self-attention computation in LLMs to address the issue.
However, they either fall short in maintaining model quality or result in high serving latency.
Drawing inspiration from advanced embedding retrieval techniques prevalent in the data management community, we consider the storage and retrieval of KVCache as a typical embedding retrieval problem.
We propose \textbf{\name}, which employs Product Quantization (PQ) to manage KVCache, maintaining model quality while ensuring low serving latency.
During the prefilling phase, we apply PQ to tokens' keys for each LLM layer and head.
During the autoregressive decoding phase, we use PQ codes and centroids to approximately identify important preceding tokens, then fetch the corresponding key-value pairs for self-attention computation.
Through meticulous design of overlapping and caching, we minimize any additional computation and communication overhead during both phases.
Extensive experiments demonstrate that \name achieves both effectiveness and efficiency, with 4.60\% score improvement over existing methods on InfiniteBench and low system latency in both prefilling and decoding.

\end{abstract}



\maketitle

\section{Introduction}

Large Language Models (LLMs), such as GPT~\cite{DBLP:journals/corr/abs-2303-08774} and Llama~\cite{DBLP:journals/corr/abs-2407-21783}, have demonstrated exceptional performance in the task of ``next token prediction'', which encompasses a broad range of applications in text understanding and generation~\cite{DBLP:journals/corr/abs-2407-21783,DBLP:journals/www/WuZQWGSQZZLXC24,DBLP:journals/corr/abs-2308-10620,DBLP:conf/icaif/LiWDC23, yang2023large}.
Beyond their extensive applications in language processing, transformer-based LLMs have also been successfully utilized in multi-modal contexts, including image analysis~\cite{DBLP:conf/iclr/DosovitskiyB0WZ21,DBLP:conf/iccv/PeeblesX23}, video processing~\cite{DBLP:journals/aiopen/RuanJ22}, software-engineering~\cite{zhang2024deep,chen2025deep,DBLP:journals/corr/abs-2308-10620}, databases~\cite{zhou2024db,zhu2024relational}, and scientific AI applications~\cite{DBLP:journals/corr/abs-2311-07361,DBLP:journals/nature/BiXZCG023}.
In recent years, the maximum input length of LLM inference has seen a substantial increase, in order to achieve better model performance through in-context learning~\cite{DBLP:journals/corr/abs-2108-07258,DBLP:conf/emnlp/MinLHALHZ22,song2024communication}, as well as to enable various complicated applications across different domains~\cite{DBLP:conf/nips/Wei0SBIXCLZ22,zheng2024open,chandra2023transformer,DBLP:conf/icml/NguyenBKGG23}.
The maximum input length has grown from 2K-4K~\cite{DBLP:journals/corr/abs-2307-09288,taori2023alpaca} to 32K~\cite{DBLP:journals/corr/abs-2310-06825,llama32k}, 128K~\cite{DBLP:journals/corr/abs-2303-08774,DBLP:journals/corr/abs-2402-10171,DBLP:journals/corr/abs-2407-21783}, and even millions of tokens~\cite{kimichat,tongyiqianwen,DBLP:journals/corr/abs-2402-08268}.

During inference, LLMs takes a sequence of tokens (also referred to as the context) as input and autoregressively generates subsequent tokens.
A typical LLM employs decoder-only self-attention modules to facilitate interaction between tokens. 
Each module calculates the ``query'', ``key'', and ``value'' representations for each token, multiplies queries and keys, and uses the softmax function to derive attention scores.
These scores are then used for weighted summation of the values.
As a decoder-only model, the query of each token can only access the keys and values of preceding tokens.
Therefore, the keys and values of previous tokens are utilized for subsequent tokens, while newly generated tokens do not influence previous ones.
The LLM inference process consists of one \textit{prefilling} phase and multiple \textit{decoding} phases.
In each phase, the LLM generates a new token.
During prefilling, the LLM processes the lengthy input and compute keys and values for all input tokens.
During each decoding phase, to avoid redundant computations, the LLM only processes the last generated token and produces its key and value.
Preceding tokens' keys and values, referred to as the Key-Value Cache (KVCache), are saved and reused by later tokens.
More details about LLM inference can be found in Section~\ref{sec:pre:llm}.

\begin{figure}[tbhp]
    \centering
    \includegraphics[width=0.4\linewidth]{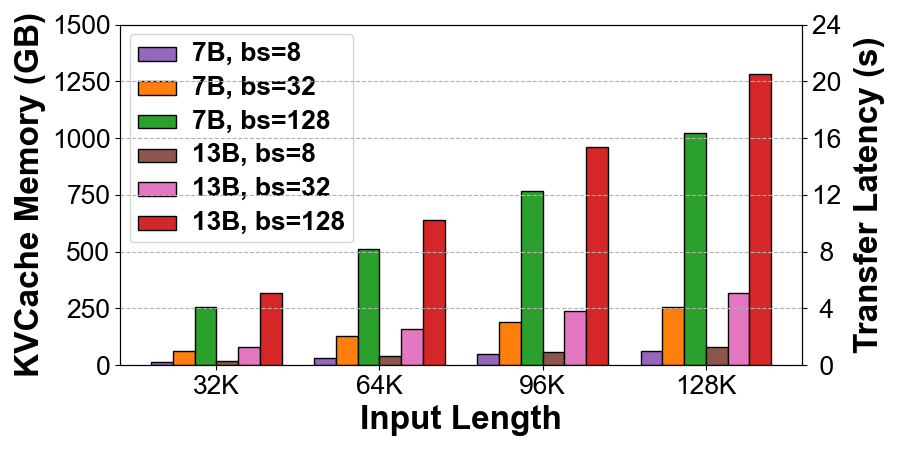}
    \caption{KVCache memory size and theoretical CPU-GPU transfer latency over PCI-e Gen 5 for varying batch sizes (bs), model sizes (7B and 13B), and sequence lengths.}
    \label{fig:intro:kvmem}
\end{figure}

As sequence lengths continue to grow in nowadays applications, the memory consumption of KVCache has increased dramatically, far exceeding the memory capacity of individual GPUs.
For instance, using a 7B model for inference on 128K-length sequences, a batch of 128 samples requires 1TB of KVCache memory as shown in Figure~\ref{fig:intro:kvmem}, which surpasses the 640GB GPU memory available in an 8-card A100 setup.
In such cases, the KVCache needs to be stored in lower-level memory hierarchies, resulting in time-consuming transfer latency and complex scheduling~\cite{DBLP:conf/icml/0007ZYLRCLRSZ23,DBLP:journals/corr/abs-2310-07240,strati2024d}, which hinders normal generation process.
Essentially, the task of managing large-scale KVCache within constrained GPU memory, while optimizing both model quality and efficiency, is a data management problem.

\begin{figure}[tbhp]

    \centering
    \includegraphics[width=0.7\linewidth]{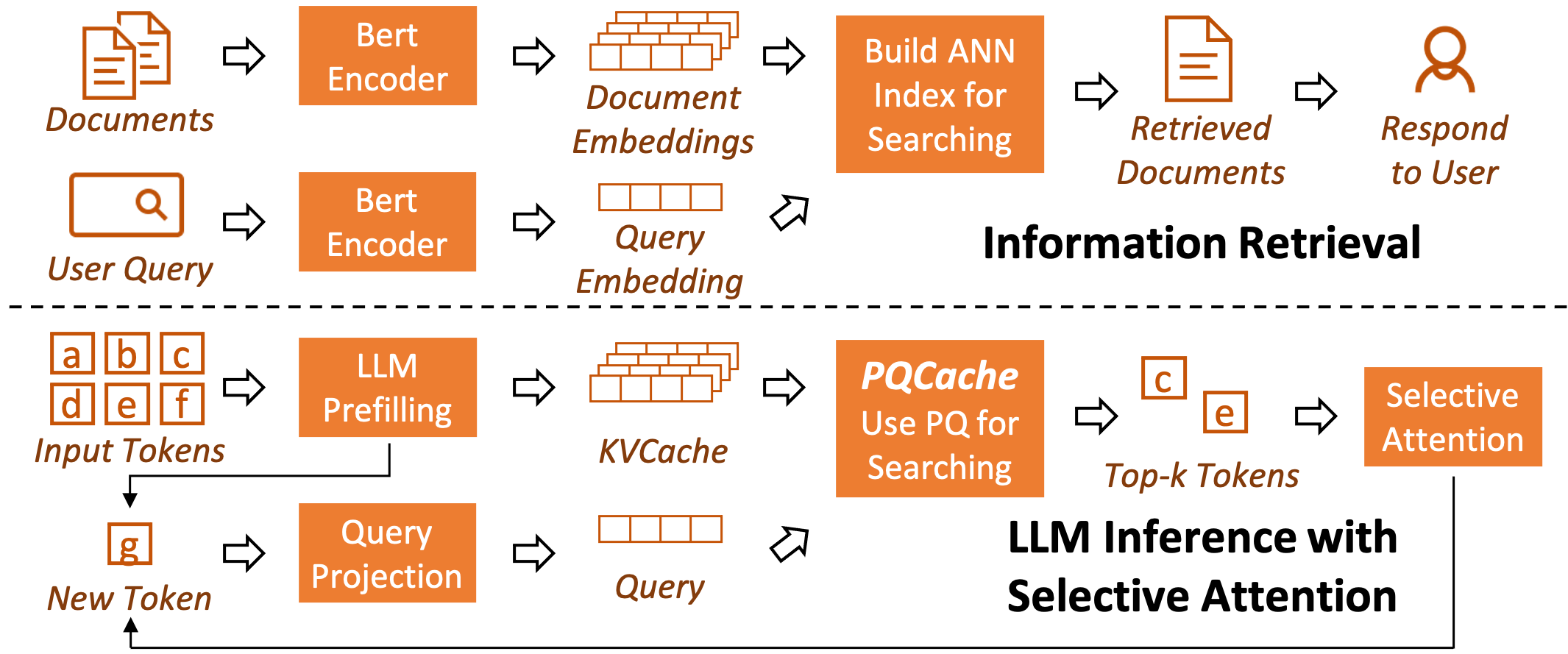}

    \caption{Comparison between information retrieval and LLM inference with selective attention.}
    \label{fig:ircompare}

\end{figure}

To address the memory challenge posed by KVCache in long-context LLM inference, a natural solution is \textit{selective attention}~\cite{miao2023towards}, which incorporates only relevant tokens for attention computation, rather than using all previous tokens.
This solution is based on the observation that, during each step of generation, certain tokens significantly influence the outcome, as their attention scores are considerably larger than others~\cite{DBLP:conf/nips/Zhang00CZC0TRBW23,DBLP:conf/nips/LiuDLWXXKS23}.
This finding is also corroborated in Section~\ref{sec:method:overview}.
Building on the concept of selective attention, numerous methods have been proposed, broadly classified into two categories: KVCache dropping~\cite{DBLP:conf/nips/Zhang00CZC0TRBW23,DBLP:conf/nips/LiuDLWXXKS23,DBLP:journals/corr/abs-2309-17453} and KVCache offloading~\cite{DBLP:journals/corr/abs-2402-04617,DBLP:journals/corr/abs-2312-04985}.
However, existing methods fail to achieve effectiveness and efficiency simultaneously, as they either rely on improper assumptions or introduce notable latency.

KVCache dropping methods discard key-value pairs deemed unimportant in earlier steps, assuming that these tokens will not affect subsequent generation.
However, the assumption often fails in real-world tasks, as tokens with initially low weights may gain importance in later steps and influence subsequent outputs~\cite{kang2024gear,DBLP:journals/corr/abs-2402-09398,DBLP:journals/corr/abs-2402-18096}.
Existing KVCache offloading methods, including SPARQ~\cite{DBLP:journals/corr/abs-2312-04985} and InfLLM~\cite{DBLP:journals/corr/abs-2402-04617}, store the entire KVCache on CPU.
For each newly generated token, they fetch relevant key-value pairs based on low-cost proxy scores.
SPARQ identifies a subset of dimensions with large magnitude in queries, fetches only these dimensions from all keys, and computes inner-product to determine the most relevant tokens.
Despite its effectiveness, SPARQ incurs substantial communication overhead that cannot be overlapped with computations.
InfLLM organizes the KVCache into blocks and uses representative tokens from each block to compute relevance.
Though the block-level space-continuity assumption improves efficiency, it does not align with real scenarios where relevant tokens are distributed discretely, leading to a significant drop in model quality.
In summary, existing methods fall short in achieving both effectiveness and efficiency for long-context LLM inference.

We observe that selective attention is essentially an information retrieval process, as it requires to find the top-$k$ relevant tokens' key-value pairs based on query-key multiplications.
Information retrieval, a prevalent research area within the database and data management domains, encompasses many well-known methods including Product Quantization~\cite{DBLP:journals/pami/JegouDS11,DBLP:conf/cvpr/GeHK013,DBLP:journals/pvldb/WangD20,DBLP:conf/cikm/LiuCC17,DBLP:journals/pvldb/AndreKS15,DBLP:conf/sigmod/Edian21,DBLP:journals/pacmmod/GaoL24}, inverted index~\cite{DBLP:conf/eccv/BaranchukBM18,DBLP:conf/nips/ChenZWLLLYW21,DBLP:journals/pacmmod/WidmoserKA24,DBLP:conf/icde/ZhouGJKLLTYZ18,DBLP:conf/sigmod/YangLFW20}, and graph-based methods~\cite{DBLP:journals/pami/MalkovY20,jayaram2019diskann,azizi2150vector,DBLP:journals/pvldb/ZhaoTHZZ23,DBLP:journals/pvldb/LuKXI21,DBLP:journals/pvldb/AziziEP23}.
As shown in Figure~\ref{fig:ircompare}, information retrieval typically involves two phases: (1) index construction, where document embedding vectors are generated by an encoder and organized into an Approximate Nearest Neighbor (ANN) index; (2) searching, which aims to efficiently retrieve the top-$k$ relevant embedding vectors from the index for a given query embedding.
Interestingly, the selective attention process in LLM inference aligns with these phases of information retrieval.
During the prefilling phase, the input tokens' KVCache is generated, with the keys serving as the embedding vectors for index construction.
Then, the decoding phase retrieves relevant tokens given the new token's query embedding vector, which corresponds to the searching phase in information retrieval.

Drawing inspiration from the field of information retrieval, we propose \textbf{\name}, a novel approach that integrates retrieval techniques into the management of KVCache to ensure both effectiveness and efficiency in long-context LLM inference.
During the prefilling phase, we generate the KVCache, store it in CPU memory, and then construct index structures.
In the decoding phase, we efficiently retrieve relevant key-value pairs using the constructed index.
Given the latency requirements of LLM inference, we do not consider methods with expensive index construction overheads, such as graph-based methods or complex inverted-index methods.
Instead, we leverage the low-cost Product Quantization (PQ) technique~\cite{DBLP:journals/pami/JegouDS11}, which partitions embeddings into sub-embeddings then conducts clustering.

The core concept of \name involves constructing PQ structures using preceding token keys and performing efficient searching to retrieve relevant key-value pairs for subsequent self-attention computations.
PQ provides a well-behaved approximation of embedding vectors (and their inner product), while consuming only a small amount of memory.
During the prefilling phase, we apply PQ to the generated keys for each layer and head, resulting in PQ codes and centroids through clustering on CPU.
At each autoregressive decoding step, we perform inner product between the partitioned query and the PQ centroids, then combine with PQ codes to obtain the approximate attention weights.
Using the approximation, we retrieve top-$k$ relevant key-value pairs from CPU memory for self-attention computation, eliminating the need to access the entire KVCache.
To further facilitate efficient LLM inference, we leverage system optimization opportunities to reduce latency.
We implement prefetching and overlapping wherever possible: KVCache offloading, PQ construction, and the fetching of PQ codes are overlapped with LLM computation.
To maximize GPU memory utilization and minimize CPU-GPU communication, we introduce a block-level cache on GPU specifically for frequently accessed key-value pairs.
Further experimental analysis show that \name improves the InfiniteBench scores by 4.60\% compared to existing methods, while maintaining low system latency.

To the best of our knowledge, this is the first work that incorporates information retrieval techniques to address the KVCache memory challenge.
Information retrieval, aimed at identifying relevant information from vast resources, is a critical component of modern web services.
As a burgeoning application, long-context LLM inference also deals with a large volume of information within the representation of the input context.
Therefore, our innovative approach of integrating retrieval techniques into LLM inference presents an intuitive and effective solution.
With the proposal of \name, we are not only addressing a current challenge but also pioneering a new domain for efficient LLM inference.
Looking ahead, we anticipate that this integration will become a standard paradigm in next-generation LLM inference, propelling the field forward and establishing a significant milestone for efficiency and effectiveness.

We summarize our contributions as follows:
\begin{itemize}[leftmargin=*,parsep=0pt,itemsep=0pt,topsep=2pt,partopsep=2pt]
    \item We effectively implement the information retrieval technique PQ into KVCache management, enhancing both model performance and efficiency of LLM inference.
    \item We propose a system-algorithm co-designed approach \name to approximately retrieve the top-$k$ relevant keys for a given query, with meticulous design of overlapping and caching.
    \item We evaluate \name through extensive experiments. It improves the InfiniteBench scores by 4.60\% compared to existing methods, while achieving low system latency.
\end{itemize}

\begin{figure}[tbhp]
    \centering
    \includegraphics[width=0.7\linewidth]{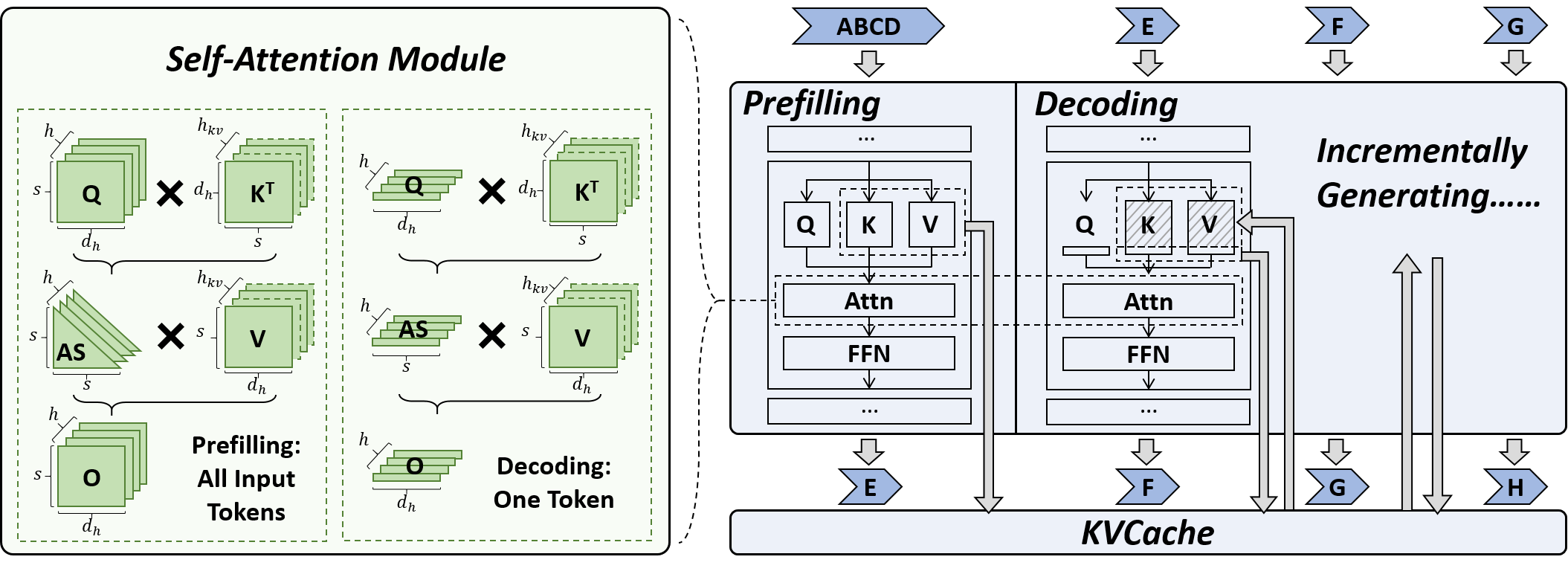}
    
    \caption{An overview of LLM inference. The left part illustrates the computation process of the self-attention module, where ``Q'', ``K'', ``V'', ``AS'', and ``O'' represent query, key, value, attention score, and output, respectively. The right part depicts the LLM inference process, consisting of the prefilling phase and the decoding phase, where ``Attn'' and ``FFN'' represent the attention layer and the feed-forward network layer, respectively. The mathematical symbols are detailed in Table~\ref{tab:notations}. }\label{fig:llm}
\end{figure}

\begin{table}[tbhp]
    \centering
    \footnotesize
    \caption{Notations. ``\#'' means ``the number of''.}
    \begin{tabular}{|c|l|c|l|}
    
     \toprule[1pt]
     \textbf{Sym.} & \textbf{Explanation} & \textbf{Sym.} & \textbf{Explanation} \\
     \midrule[0.8pt]
     $n$ & Batch size.  & $m$ & \# partitions in PQ. \\
     $s$ & Current sequence length. & $b$ & \# bits for PQ codes. \\
     $d$ & Hidden states dimension.  & $d_{m}$ & Dimension of each partition. \\
     $h_{(kv)}$ & \# heads (for keys and values). & $k$ & \# tokens in selective attention. \\
     $d_{h}$ & Dimension of each head. & $T$ & \# K-Means iterations. \\
     \bottomrule[1pt]
     \end{tabular}
    \label{tab:notations}
    
    \end{table}

\section{Preliminary}

In this section, we introduce fundamental concepts related to LLM, PQ, and the memory hierarchy.

\subsection{Large Language Model Inference}\label{sec:pre:llm}

An overview of LLM inference is depicted in Figure~\ref{fig:llm}.
An LLM consists of a vocabulary embedding for input, a stack of transformer layers, and a token classifier for output.
The self-attention module, which is a crucial component of a transformer layer, facilitates interaction and information aggregation among different tokens.
Following the notations in Table~\ref{tab:notations}, each transformer layer receives an input of shape $(n, s, d)$.
The input is separately projected and transposed for query, key, value, resulting in the same shape of $(n, h, s, d_h)$, where it usually holds that $d = h * d_h$.
Different heads are expected to capture different semantic information.
The attention mechanism multiplies the queries and keys, applies a lower-triangular causal mask to restrict queries to preceding keys only, and performs softmax to obtain the attention scores of shape $(n, h, s, s)$.
The attention scores are then used to weighted-sum the values, yielding an output of shape $(n, h, s, d_h)$.
In long-context inference, the $O(s^2)$ space complexity associated with operations among queries, keys, and values is impractical.
Consequently, modern LLMs typically employ FlashAttention~\cite{DBLP:conf/nips/DaoFERR22,DBLP:conf/iclr/Dao24}, which utilizes tiled matrix multiplication and softmax, to only use $O(s)$ space complexity for attention computation.
It's worth noting that, the time complexity of attention computation remains $O(s^2)$.
The output from attention is later reshaped into $(n, s, d)$  for the subsequent fully-connected layers, also known as Feed-Forward Network (FFN).
To alleviate memory and computation burden, Grouped-Query Attention (GQA) is often applied.
It employs a smaller number of heads $h_{kv}$ for keys and values, resulting in their shape being $(n, h_{kv}, s, d_h)$.
In this setup, each key-value pair corresponds to multiple queries.

During LLM inference, each execution of the model generates a new token, following an autoregressive manner.
The first traversal of the LLM is called ``prefilling'', and subsequent traversals are referred to as ``decoding'', as shown in Figure~\ref{fig:llm}.
In the prefilling phase, the self-attention module computes the queries, keys, and values for all input tokens, and stores the key-value pairs as KVCache for later use.
Thanks to the causal mask, later tokens do not affect earlier tokens, so in the autoregressive decoding phase we only need to compute the query, key, value for the last generated token.
This process leverages previous keys and values from the KVCache, and computes an attention score of shape $(n, h, 1, s)$.
Concurrently, the newly generated key and value are added to the KVCache. 
As a result, the memory consumption of KVCache scales linearly with the sequence length, leading to a memory bottleneck in scenarios involving long-context input and output.

\subsection{Product Quantization}\label{sec:pre:pq}

\begin{figure}[tbhp]
    \centering
    \includegraphics[width=0.5\linewidth]{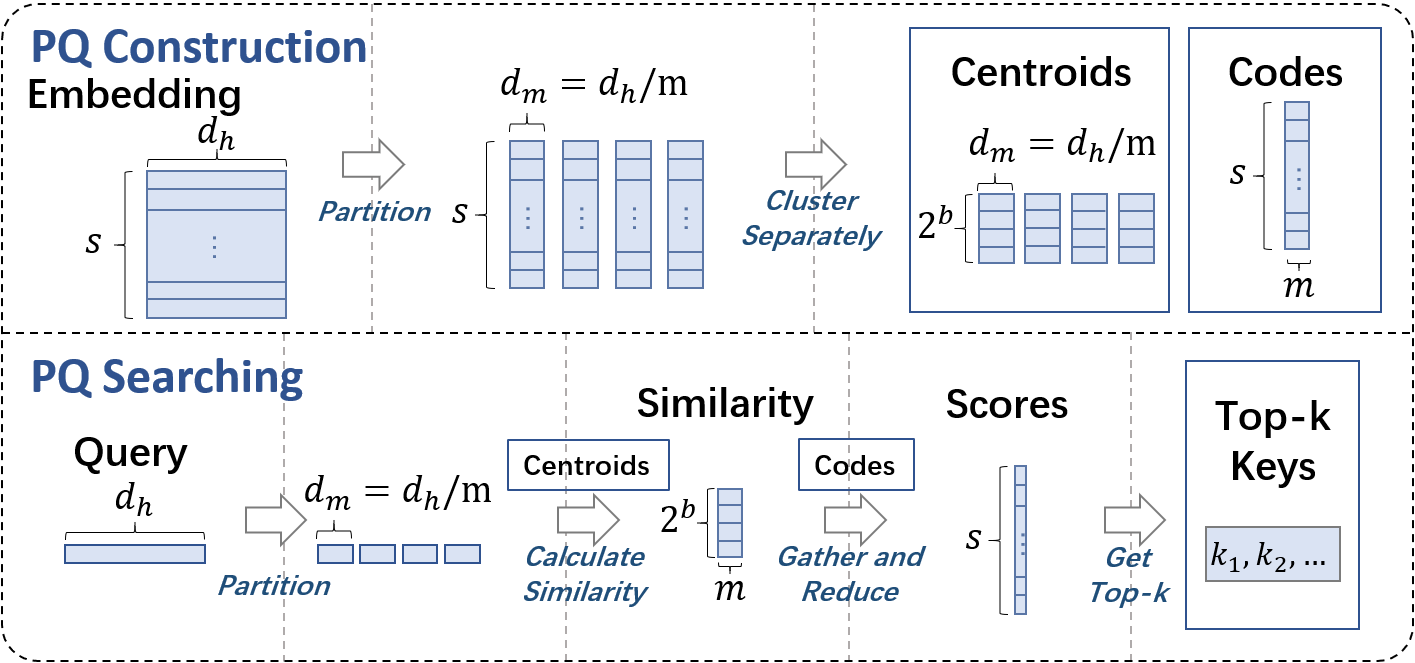}

    \caption{An overview of PQ construction and searching.}
    \label{fig:pq}

\end{figure}

PQ~\cite{DBLP:journals/pami/JegouDS11} was proposed to facilitate efficient Approximate Nearest Neighbor Search (ANNS), retrieving relevant embedding vectors from a large pool of candidates given a query embedding vector.
As shown in Figure~\ref{fig:pq}, PQ involves two phases: index construction and searching. 
During construction, PQ divides each candidate embedding into $m$ partitions, essentially decomposing the original embedding space into $m$ separate sub-spaces.
Each sub-space undergoes K-Means clustering to group the sub-embeddings, yielding $2^b$ centroids.
Each embedding is then assigned $m$ codes, each with $b$ bits, indicating the centroids it belongs to. 
These compact PQ codes enable the reconstruction of approximate embeddings using centroids, significantly reducing memory requirements.
During searching, the query embedding is divided into $m$ partitions. 
Similarity is computes with the centroids within each partition, and aggregated through PQ codes, bypassing the need for full similarity calculations with every embedding.

PQ has a profound impact on ANNS, with its principles integrated into various efficient ANNS methods~\cite{DBLP:journals/tbd/JohnsonDJ21,jayaram2019diskann,DBLP:conf/eccv/BaranchukBM18}.
While PQ was initially designed together with IVF (inverted file system)~\cite{DBLP:journals/pami/JegouDS11} for ANNS, PQ and IVF are independent techniques and are often applied separately~\cite{DBLP:conf/cvpr/GeHK013,DBLP:journals/tbd/JohnsonDJ21,DBLP:conf/sigir/XiaoLHZLGCYSSX22}.
PQ's variants are also applied in various learning tasks~\cite{DBLP:journals/corr/abs-2309-16354,DBLP:conf/nips/OordVK17,DBLP:conf/nips/ZhangWCCZMHDMWP23} to achieve effective compression and efficient computation.
In this paper, we utilize PQ in long context LLM inference to ensure both effectiveness and efficiency.
For potential future extensions involving IVF or other ANNS tecniques, please refer to Section~\ref{sec:discussion} for further discussion.

\subsection{GPU-CPU Memory Hierarchy}\label{sec:pre:mem}

Modern deep learning tasks heavily rely on GPUs for executing compute-intensive operations.
The GPU-CPU structure forms a typical memory hierarchy: the more expensive GPU memory offers faster memory I/O speeds for computation, while the CPU memory, connected via PCI-e or NVLink, provides lower bandwidth.
This memory hierarchy and the GPU-CPU interconnect is common in modern GPU-equipped servers utilized in cloud environments~\cite{DBLP:conf/nsdi/WengXYWWHLZLD22,DBLP:conf/nsdi/WuZBL023}, data centers~\cite{DBLP:journals/csur/YeGHSWLZW24,DBLP:conf/nsdi/Hu0WWZC0L0L0024}, and edge devices~\cite{DBLP:journals/comsur/WangHLNYC20,DBLP:journals/tpds/LiuZLWWFZD22}.
This hardware setting allows us to leverage available CPU memory for better efficiency when GPU memory is insufficient.
As model parameters increase and the demand for intermediate results storage (such as KVCache) grows, CPUs are often employed to share the memory load.
Numerous research studies in machine learning systems propose offloading certain model parameters or activations to the CPU memory~\cite{DBLP:conf/micro/RhuGCZK16,DBLP:conf/ipps/BGK19,DBLP:conf/sc/RajbhandariRRSH21,DBLP:conf/usenix/0015RARYZ0H21,DBLP:conf/icde/NieMYC22,DBLP:conf/icml/0007ZYLRCLRSZ23,DBLP:journals/corr/abs-2402-04617,accelerate}, thereby enhancing the overall performance of GPU-centric deep learning tasks.
Similar to these studies addressing memory bottlenecks in deep learning models, we also treat GPU memory as a premium, high-bandwidth resource and consider CPU memory a secondary tier.
The primary challenge in this context is to effectively schedule memory I/O (or say GPU-CPU communication) in conjunction with GPU computation to efficiently hide the associated overhead.
It is worth noting that, although we currently implement \name within the GPU-CPU memory hierarchy, it can also be extended to incorporate disk storage — thereby adding another level to the memory hierarchy — to accommodate environments with limited resources in the future.

\section{\name}
\label{sec:method}

\begin{figure}[tbhp]
    \centering
    \includegraphics[width=0.7\linewidth]{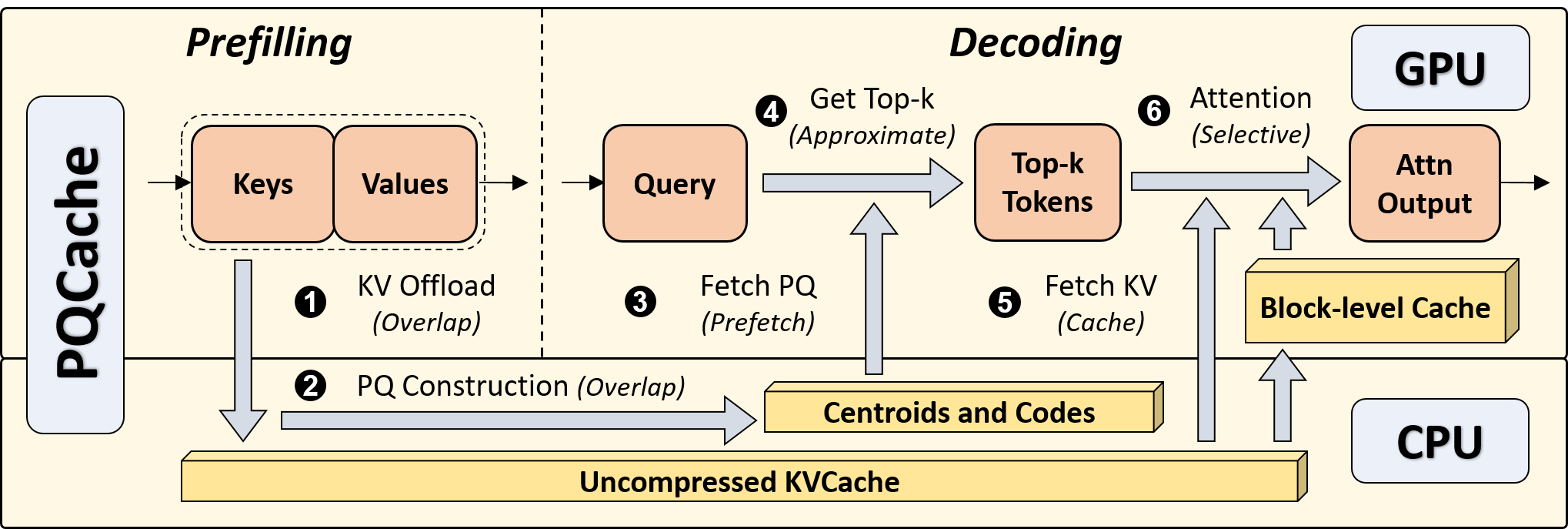}

    \caption{An overview of \name. For simplicity, we only illustrate the process for a single transformer layer.}
    \label{fig:pqcache}

\end{figure}

In this section, we introduce \name, a novel system-algorithm co-designed method to enable effective and efficient long context LLM inference with large-scale KVCache.
Figure~\ref{fig:pqcache} provides an overview of \name, where the KVCache from the prefilling phase is first offloaded to CPU, compressed using PQ, then fetched on demand during the decoding phase.

\subsection{Overview}
\label{sec:method:overview}

\begin{figure}[tbhp]
    \centering
    \subfigure[Layer 3, head 25.]{
    \scalebox{0.2}{
    \includegraphics[width=\linewidth]{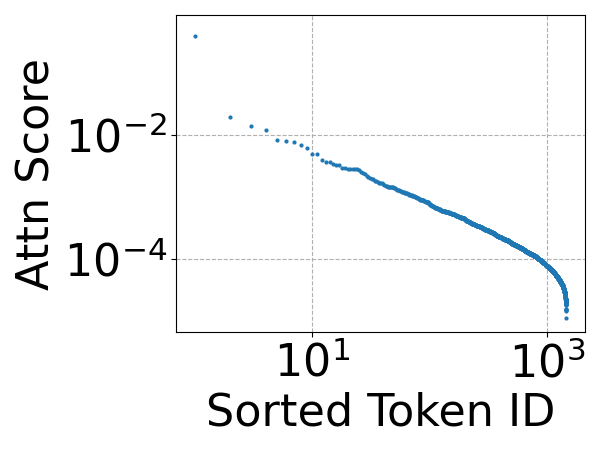}
    }
    }
    \subfigure[Layer 11, head 15.]{
    \scalebox{0.2}{
    \includegraphics[width=\linewidth]{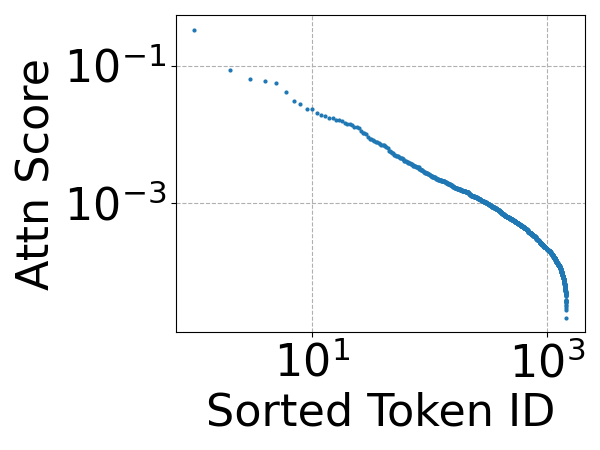}
    }
    }
    \subfigure[Layer 20, head 27.]{
    \scalebox{0.2}{
    \includegraphics[width=\linewidth]{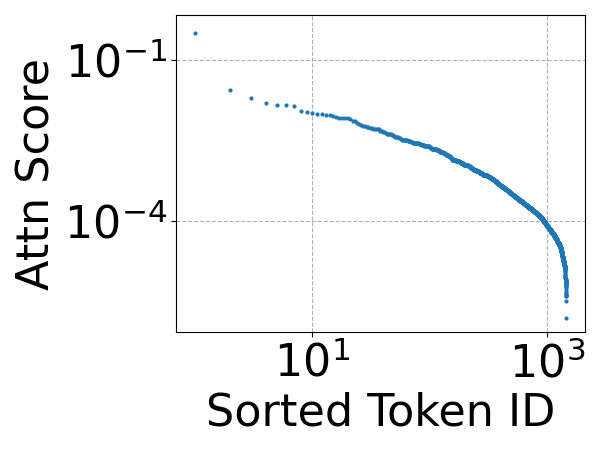}
    }
    }
    \subfigure[Layer 21, head 16.]{
    \scalebox{0.2}{
    \includegraphics[width=\linewidth]{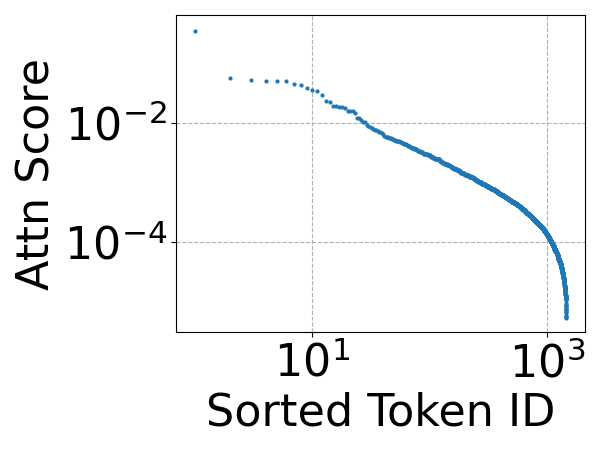}
    }
    }
    \caption{Distributions of attention scores.}
    \label{fig:xsum_attn_dist}
\end{figure}

We design \name to reserve all the KVCache in CPU, and selectively fetch relevant key-value pairs for self-attention computation.
In long context inference scenario, the entire KVCache is too large for both attention computation and I/O communication within the memory hierarchy.
Therefore, a common technique is to only perform attention on a subset of the key-value pairs, a process known as ``selective attention''.
According to previous research~\cite{DBLP:conf/nips/Zhang00CZC0TRBW23,DBLP:conf/nips/LiuDLWXXKS23,DBLP:journals/corr/abs-2402-04617,DBLP:journals/corr/abs-2312-04985,DBLP:journals/corr/abs-2402-18096,DBLP:journals/corr/abs-2403-09054,DBLP:journals/corr/abs-2310-01801,wang2024squeezeattention,DBLP:journals/corr/abs-2402-06262}, attention score is a proper metric to measure the importance or relevance of previous tokens.
As shown in Figure~\ref{fig:xsum_attn_dist}, we plot the attention score distributions at several randomly-selected positions on an example from a common summarization dataset XSUM~\cite{DBLP:conf/emnlp/NarayanCL18}.
The attention scores generally follow powerlaw distributions, indicating that a small part of tokens are more important than most other tokens.
Therefore, we can only include those tokens with large scores for self-attention computation.
Following prior works~\cite{DBLP:journals/corr/abs-2309-17453,DBLP:journals/corr/abs-2308-16137,DBLP:conf/nips/Zhang00CZC0TRBW23}, we also include initial tokens and the most recent tokens (called local tokens) in attention computation.

As detailed in Section~\ref{sec:pre:llm}, attention scores are calculated using a softmax function applied to the product of the current query and preceding keys.
The procedure of identifying the top-$k$ keys with the highest scores fundamentally constitutes an Approximate Nearest Neighbor Search (ANNS) operation.
Therefore, we try to leverage embedding retrieval techniques to enable effective selective attention and address the KVCache memory issue.
Based on the observations above, we design \name, which offloads all the KVCache to CPU, and fetch only relevant tokens' key-values pairs during the decoding phase.
Calculating exact attention scores of all previous tokens involves costly I/O communication, which is unacceptable in long context LLM inference.
Inspired by ANNS in information retrieval~\cite{DBLP:journals/tbd/JohnsonDJ21,jayaram2019diskann,DBLP:conf/eccv/BaranchukBM18}, we leverage the light-weight Product Quantization (PQ) method~\cite{DBLP:journals/pami/JegouDS11}, which compress the vectors by partitioning and K-Means clustering.
Though there are other ANNS methods (e.g. graph-based methods~\cite{DBLP:journals/pami/MalkovY20,DBLP:journals/pvldb/FuXWC19,jayaram2019diskann}) that can achieve better recall performance, they suffer from a computationally expensive construction process which may hinder LLM inference.

In \name, we construct PQ during the prefilling phase and utilize PQ during the decoding phase.
We use tensors to manage all PQ-related data structures, which are natively supported by deep learning frameworks including PyTorch~\cite{DBLP:conf/nips/PaszkeGMLBCKLGA19}, TensorFlow~\cite{DBLP:conf/osdi/AbadiBCCDDDGIIK16}, and Hetu~\cite{miao2023hetu}.
The following is a detailed explanation of how \name operates, as illustrated in Figure~\ref{fig:pqcache}.

Step \ding{182}: During the prefilling phase, we calculate all the input tokens' keys and values at each layer.
Both the keys and values have the shape of $(n, h_{kv}, s, d_h)$.
They are then asynchronously offloaded to CPU , a process that can overlap with subsequent computations.

Step \ding{183}: We perform PQ construction on CPU.
We omit the batch size and head dimensions for simplicity, and get per-head key vector of shape $(s, d_h)$.
We then divide the dimension $d_h$ into $m$ sub-spaces with dimension $d_m$.
For partitioned vectors $(m, s, d_m)$, where $d_m=d_h/m$, we conduct K-Means clustering for each group separately, resulting in centroids of shape $(m, 2^b, d_m)$ and PQ codes of shape $(s, m)$.
Each PQ code, which indicates the cluster to which the vector belongs, requires only $b$ bits for storage.

Step \ding{184}: During the decoding phase, while computing the previous transformer layer, we pre-fetch PQ centroids and codes.
While the PQ codes are linear with sequence length, the centroids are of small volume and can be kept on GPU throughout the inference.

Step \ding{185}: We perform PQ search on GPU: first conducting matrix multiplication between the query and the PQ centroids, then aggregating the result with PQ codes to obtain the approximate scores for all tokens.
We can then identify the top-$k$ relevant tokens using the approximate scores.

Step \ding{186}: Using the PQ scores, we fetch the approximate top-$k$ tokens' key-value pairs from CPU, or from a GPU cache which we will discuss later in Section~\ref{sec:method:decode}.

Step \ding{187}: The selective self-attention computation continues with the retrieved tokens.

Unlike typical embedding retrieval tasks, in LLM inference, newly generated keys and values are added to the KVCache.
These tokens are initially considered as local tokens and reserved in GPU.
When they are evicted from the sliding window of local tokens, they are assigned PQ codes based on their nearest centroids.

\subsection{Complexity Analysis}
\label{sec:method:complexity}

In this section, we analyze the time and space complexity in \name-equipped LLM inference.
During prefilling, the attention computation remains unmodified, thus the time and memory complexity stay the same.
Concretely, the matrix multiplications in attention and FFN have the tensor shapes of $(h, s, d_h)\times (h, d_h, s) \rightarrow (h, s, s)$, $(h, s, s)\times (h, s, d_h) \rightarrow (h, s, d_h)$, and $(s, d)\times (d, d) \rightarrow (s, d)$, yielding a time complexity of $O(s^2d/h + sd^2)$.
In current long-context training and inference of LLMs, FlashAttention~\cite{DBLP:conf/nips/DaoFERR22,DBLP:conf/iclr/Dao24} has become the de facto implementation of attention computation.
It implements tiled multiplication and softmax, resulting in a space complexity of $O(sd)$.
While the details of FlashAttention are not the focus of our paper, it's worth noting that in our experiments, FlashAttention is utilized by baselines and \name by default, except for H2O due to compatibility issues.
Using \name, the additional K-Means clustering process has an average complexity of $O(s h_{kv} m d_m 2^b T)$, where $T$ is the number of K-Means iterations.
Its complexity is linear with the input sequence length.
We leverage idle CPU resources to perform K-Means, allowing the clustering to overlap with GPU computation, which will be detailed in Section~\ref{sec:method:prefill}.

During the decoding phase, the matrix multiplications in the original LLM have the tensor shapes of $(h, 1, d_h)\times (h, d_h, s) \rightarrow (h, 1, s)$, $(h, 1, s)\times (h, s, d_h) \rightarrow (h, 1, d_h)$, and $(1, d)\times (d, d) \rightarrow (1, d)$, leading to a time complextiy of $O(sd + d^2)$.
Using \name, we first multiply the query with PQ centroids, which has shapes of $(h, m, 1, d_m)\times (h, m, d_m, 2^b) \rightarrow (h, m, 1, 2^b)$, then gather and reduce using PQ codes, which has shapes of $(h_{kv}, s, m) (h_{kv}, m, 1) \rightarrow (h_{kv}, s)$.
The PQ-related process has a time complexity of $O(2^b d^2/(hm) + h_{kv}ms)$.
Finding the top-$k$ largest scores has a time complexity of $O(h_{kv}s)$, since PyTorch tends to use a radix-sort-like algorithm to determine the $k$-th large element.
Computing the selective attention and the subsequent FFN has a time complexity of $O(kd + d^2)$.
Therefore, the overall time complexity is $O(2^b d^2/(hm) + h_{kv}ms + kd + d^2)$.
The memory complexity of PQ centroids and codes is $O(h_{kv}ms + h_{kv}2^b\cdot d_h)$.
In long-context scenarios, we focus on the multiplers of $s$.
For time complexity, current multipler $h_{kv}m$ is much smaller than the original $d$, considering $h_{kv}m \ll d$ (e.g. $h_{kv}$=8, $m$=2, $d$=4096 in 7B model).
For space complexity, the small size of $h_{kv}m$ allows the communication of PQ structures to overlap with GPU computation.
Consequently, \name enables more efficient decoding.

\begin{figure}[tbhp]
    \centering
    \subfigure[Prefilling.]{
    \includegraphics[width=0.7\linewidth]{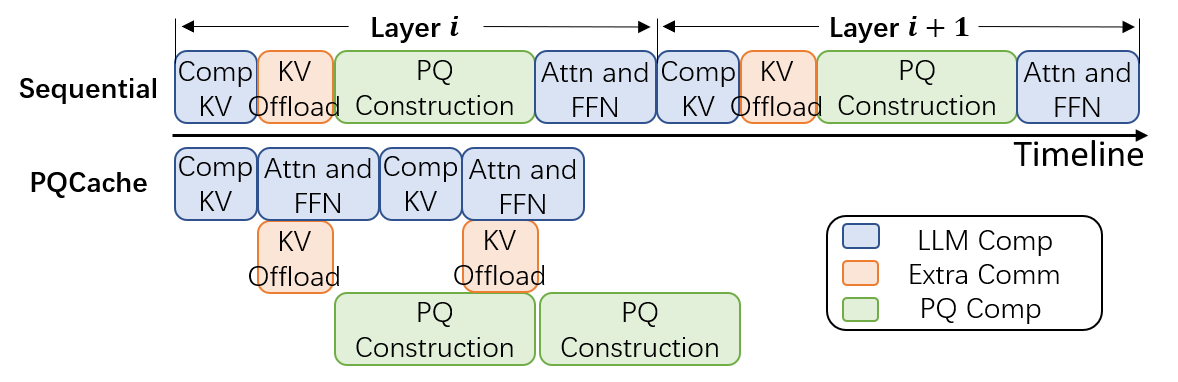}
    \label{fig:timeslot:prefill}
    }
    \subfigure[Decoding.]{
    \includegraphics[width=0.7\linewidth]{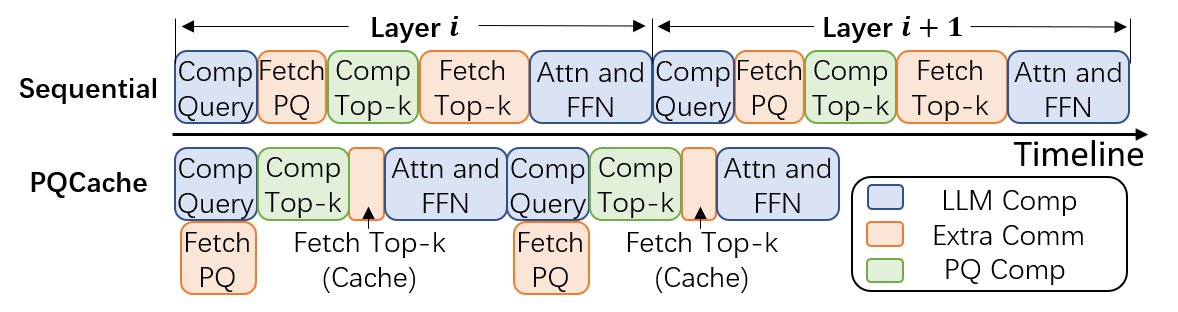}
    \label{fig:timeslot:decode}
    }
    \caption{PQCache v.s. sequential scheduling.}
    \label{fig:timeslot}
    
\end{figure}

To facilitate efficient long context LLM inference, the design goal of \name is to provide overhead-agnostic service.
Figure~\ref{fig:timeslot} illustrates the computation and communication involved in the \name-enabled LLM inference, covering both the prefilling and decoding phases.
The original LLM computation is filled with blue color, while the computation and the communication introduced by \name are filled with green and red colors, which can be divided into four parts: 
(1) KVCache offloading and PQ structure fetching;
(2) PQ construction using K-Means clustering;
(3) Approximate top-$k$ computation;
(4) The fetch process of top-$k$ relevant tokens' key-value pairs.
As shown in Figure~\ref{fig:timeslot}, except for the low-cost top-$k$ approximation, we employ distinct system design to eliminate these computation or communication overhead.
The system design is detailed in the following sections.

\subsection{Prefilling Phase}
\label{sec:method:prefill}

\begin{algorithm}
    \caption{\name prefilling phase}\label{alg:prefill}
    \Input{Tokenized query $X$}
    \Output{Output token; KVCache; PQ structures}
    $ProcComm = ProcPQ = KVCache = PQ = \{\}$;

    $L_x = X.length()$;

    \nonl // Prefilling computation on GPU 

    \For{$i\gets0$ \KwTo $L-1$}{

    $Q, K, V = Projection_i(X)$;

    $ProcComm.append(AsyncGPU2CPU(K, V))$;

    $X = AttnFFN_i(Q, K, V)$;

    }

    $Token = Classifier(X)$; 

    \nonl // Launch PQ construction on CPU

    \For{$i\gets0$ \KwTo $L-1$}{

    $K, V = ProcComm[i].sync()$;

    $KVCache.append((K, V))$; 

    $ProcPQ.append(AsyncPQConstruct(K, L_x))$;

    }

    \nonl // Wait for PQ construction in the next decoding phase

    \For{$i\gets0$ \KwTo $L-1$}{

    $Centroids, Codes = ProcPQ[i].sync()$;

    $PQ.append((Centroids, Codes))$; 

    \nonl // Some decoding logic here

    }

\Return Token, KVCache, PQ
\end{algorithm}

\begin{figure}

    \centering
    \includegraphics[width=0.4\linewidth]{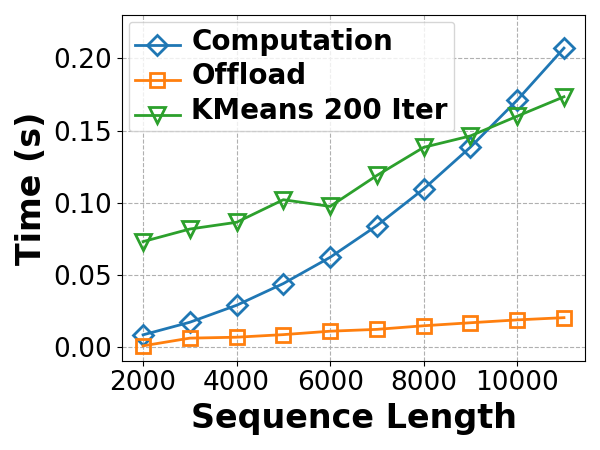}

\caption{The execution time of one-layer transformer computation, offloading, and clustering at the prefilling phase.}
    \label{fig:timing}
\end{figure}

During the prefilling phase, GPU computations, GPU-to-CPU offload communications, and PQ construction can all be executed concurrently.
The process is depicted in Figure~\ref{fig:timeslot:prefill} and Algorithm~\ref{alg:prefill}.
Lines 3-8 illustrate the GPU computations.
Within each layer, the input is initially projected into queries, keys, and values, respectively.
The keys and values are then asynchronously transferred to the CPU.
Meanwhile, the GPU continues with computations for the attention and feed-forward network operations.
Following the transformer layers, the classifier layer performs matrix multiplication and applies a softmax function to predict the next token.
Lines 9-13 and lines 14-17 demonstrate the initiation and waiting period of PQ construction, respectively.
Upon acquiring the keys and values, a separate process on CPU launches multiple clustering processes for PQ construction.
Given the distinct latent spaces of different heads in transformers $h_{kv}$ and different groups in PQ $m$, these clustering processes, totaling $h_{kv}m$ per layer per sample, can operate in parallel.
To enable on-demand synchronization, we wait for PQ construction to complete at the same transformer layer of the next token decoding phase.
Figure~\ref{fig:timing} shows the time usage of a 7B LLM single-layer transformer computation on an NVIDIA RTX 4090 GPU, GPU-to-CPU communication via PCI-e 1.0 (x16), and PQ construction (K-Means clustering processes) on two Intel(R) Xeon(R) Gold 6330 CPUs, with the input sequence lengths varying.
Given that the attention computation time scales quadratically with sequence length, while both the communication time and the K-Means clustering time scale linearly, the computation can entirely encompass the other two operations when the sequence length is sufficiently long.
In such cases, we can fully utilize the idle CPU and PCI-e resources.
However, when the sequence length is insufficient, the duration of the K-Means clustering may exceed that of the GPU computation.
This issue may become increasingly pronounced as GPU computation capability advances over the years.

To address this issue and ensure that the clustering can be overlapped by GPU computation, we propose an adaptive K-Means method that limits the maximum number of iterations for clustering.
Given the complexity of K-Means as discussed in Section~\ref{sec:method:complexity}, we fit a linear curve of K-Means clustering time as follows:
\begin{equation}
Time_{clus} = \alpha_1 + \beta_1\cdot s  T.
\end{equation}
Here $s$ is the sequence length, and $T$ is the number of iterations.
Meanwhile, the GPU computation time can be fitted as a quadratical relation:
\begin{equation}
Time_{comp} = \alpha_2 + \beta_2\cdot s + \gamma_2\cdot s^2,
\end{equation}
The maximum number of iterations, denoted as $T_{max}$, that ensures clustering does not interfere computation should satisfy $Time_{clus} = Time_{comp}$, which yields:
\begin{equation}\label{eq:tmax}
T_{max} = \frac{\gamma_2\cdot s^2 + \beta_2\cdot s + \alpha_2 - \alpha_1}{\beta_1\cdot s}.
\end{equation}
In practice, we also clip $T_{max}$ to ensure that the number of iterations is neither too small nor too large.
For any given models and devices, we can profile time of single-layer computation and clustering time using several input sequence lengths, then employ a simple regression technique to fit the coefficients $\alpha_1, \beta_1, \alpha_2, \beta_2, \gamma_2$.
By modeling the relationship between computation time and sequence length, we can determine the maximum number of K-Means iterations as Eq~\ref{eq:tmax}.
In Section~\ref{sec:expr:kmeans}, we empirically study the trade-off between the efficiency and model quality of different clustering iterations.

\begin{algorithm}
    \caption{\name decoding phase}\label{alg:decode}
    \Input{Last generated token $X$}
    \State{InitKV; MidKV; LocalKV; PQ codes; PQ centroids}
    \Output{Output token}

    \nonl // First pre-fetch first layer's PQ codes

    $Proc = AsyncCPU2GPU(PQCodes[0])$;

    \For{$i\gets0$ \KwTo $L-1$}{

    \nonl // Evict previous token from local tokens

    $EvictK, EvictV = LocalKV.pop(0)$;

    $NewCode = PQArgNearest(EvictK)$;

    \nonl // On CPU, MidKV and PQ will be updated

    $AsyncGPU2CPU((NewCode, EvictK, EvictV))$;

    \nonl // Get Q, K, V for last token

    $Q, K, V = Proj_i(X)$;

    \nonl // Get KV for current attention

    $LocalKV.append((K, V))$;

    $Codes = Proc.sync()$;

    $Centroids = PQCentroids[i]$;

    \If{$i \ne L-1$}{

    \nonl // Pre-fetch next layer's PQ codes

    $Proc = AsyncCPU2GPU(PQCodes[i+1])$;

    }

    $Codes.append(NewCode)$;

    $TopkToken = PQSearch(X, Centroids, Codes)$;

    $TopkKV = SyncFetchKV(TopkToken)$;

    $AllKV = InitKV + TopkKV + LocalKV$;

    \nonl // Continue attention and FFN computation

    $X = AttnFFN_i(Q, AllKV)$;

    }

    $Token = Classifier(X)$; 

    \Return Token

\end{algorithm}

\subsection{Decoding Phase}
\label{sec:method:decode}

Algorithm~\ref{alg:decode} outlines the decoding phase of \name.
During this phase, the constructed PQ structure is utilized by the attention module in each layer.
The PQ centroids are directly stored on GPU, since they consume a fixed and negligible amount of memory (e.g., only 16M for $m=2$, $b=6$, 7B GQA LLM), and are independent of sequence length.
As the preceding layer's computation progresses, the PQ codes for the next layer can be pre-fetched in parallel, as indicated in line 1 and lines 9-11.
Given the significantly smaller size of the PQ codes compared to original keys, their communication can overlap with the decoding phase computation.

We partition the entire KVCache into three segments: initial tokens, middle tokens, and local tokens.
The key-value pairs of initial and local tokens (the most recently generated tokens) are directly involved in attention computation and are therefore stored on GPU.
Conversely, the key-value pairs of middle tokens are stored on CPU.
In each transformer layer, local tokens first evict the earliest token, generate its PQ codes, and asynchronously offload its key and value to the CPU, as depicted in lines 3-5.
The newly generated key and value are then added to the local tokens, as shown in lines 6-7.
Upon receiving the PQ codes from the CPU in line 8, a PQ search is conducted to identify the top-$k$ tokens in lines 12-13.
After fetching these tokens' keys and values from CPU in line 14, we obtain all keys and values for attention in line 15, and the computation continues in line 16.

The only communication that cannot be overlapped is the retrieval of the top-$k$ relevant tokens, as it relies on the preceding PQ search.
Inspired by previous research~\cite{DBLP:conf/nips/Zhang00CZC0TRBW23,DBLP:conf/nips/LiuDLWXXKS23,DBLP:journals/corr/abs-2402-04617}, there are certain pivotal tokens that are consistently important during inference.
To accommodate these tokens, we design a block-level GPU cache, employing either a Least Frequently Used (LFU) or Least Recently Used (LRU) eviction policy.
On identifying the top-$k$ tokens, we first check the cache for their availability.
For tokens already in the cache, we directly retrieve their keys and values.
For those not in the cache, we fetch them from CPU.
After retrieving keys and values, we asynchronously update the cache, adding and evicting tokens according to the eviction policy.
This update process, only involves PCI-e communication, does not interfere with GPU computation.
To minimize cache-lookup overhead, we leverage block-level granularity over token-level granularity. 
We partition all tokens into blocks and store frequently-used blocks in the GPU cache.
During each retrieval, we update the cache using the top-$k_{cache}$ blocks, which contain the most top-$k$ tokens.
Here $k_{cache}$ refers to the number of blocks, while $k$ specifies the number of tokens.
Experimental results in Section~\ref{sec:latency:cache} demonstrate the effectiveness of the GPU cache in reducing latency.

\section{Experiments}
\label{sec:expr}

In this section, we conduct experiments and compare \name with existing methods.
We experimentally show that \name achieves both effectiveness and efficiency.

\subsection{Experimental Setup}

\subsubsection{\textbf{Models}}

We conduct experiments using two representative open-source LLMs: Llama-3.1-8B~\cite{DBLP:journals/corr/abs-2407-21783} and Mistral-7B-Instruct-v0.2~\cite{DBLP:journals/corr/abs-2310-06825}, which support context lengths of 128K and 32K, respectively.
Both models employ GQA and share similar architectures.
Following common practice, we use FP16 for model inference.

\subsubsection{\textbf{Tasks}}

We evaluate \name on two multitask benchmarks for long context inference: LongBench~\cite{DBLP:journals/corr/abs-2308-14508} and InfiniteBench~\cite{zhang-etal-2024-bench}.
These benchmarks are widely evaluated in related research works~\cite{DBLP:journals/corr/abs-2310-01801,DBLP:journals/corr/abs-2312-04985,DBLP:journals/corr/abs-2402-04617,wang2024squeezeattention,DBLP:journals/corr/abs-2404-14469,DBLP:journals/corr/abs-2406-02069,DBLP:journals/corr/abs-2407-15891}.
The evaluation tasks encompass question answering, summarization, few-shot learning, multiple choice, and retrieval.
Detailed meta information such as sample lengths, task descriptions, and metrics is available on LongBench\footnote{https://github.com/THUDM/LongBench/tree/main/LongBench} and InfiniteBench\footnote{https://github.com/OpenBMB/InfiniteBench} websites.
Tasks in LongBench typically average 10k in length, while those in InfiniteBench average 100k, which is significantly longer.
Each task employs a specific evaluation metric, such as accuracy, Rouge-L, or F1-score, which we simply call ``score''.
Consistent with previous work, we report the scores of the tasks and their average.

We also experiment on two additional tasks: the Needle-in-a-Haystack~\cite{needleinahaystack} and the GSM8k Chain-of-Thought (CoT) reasoning~\cite{DBLP:conf/nips/Wei0SBIXCLZ22}.
The Needle-in-a-Haystack test evaluates the in-context retrieval ability of long-context LLMs by asking the model to retrieve a random fact or statement placed from a lengthy document.
This test is conducted on documents with various lengths to determine the model's retrieval accuracy.
In our experiments, we consider up to 128K document length.
GSM8k is a math reasoning dataset containing 8k high quality diverse grade school math problems.
Its CoT variant is a complex reasoning task that require model to attend to extensive contextual details for accurate answers, with an average input length of 3.7k.
The accuracy metric measures if the model can correctly answer the questions.

\subsubsection{\textbf{Baselines}}
\label{sec:expr:settings:baselines}

We compare with the following baselines: 
\begin{itemize}[leftmargin=*,parsep=0pt,itemsep=0pt,topsep=2pt,partopsep=2pt]
    \item \textbf{H2O}~\cite{DBLP:conf/nips/Zhang00CZC0TRBW23} uses accumulative attention scores to determine which key-values pairs to retain. 
    \item \textbf{SnapKV}~\cite{DBLP:journals/corr/abs-2404-14469} analyzes the attention scores from the prompt's last segment to identify important tokens and employs pooling to preserve their surrounding information.
    \item \textbf{PyramidKV}~\cite{DBLP:journals/corr/abs-2406-02069} improves over SnapKV by allocating more budget to lower layers and less to higher layers. 
    \item \textbf{SPARQ}~\cite{DBLP:journals/corr/abs-2312-04985} selects dimensions with the largest magnitudes in query tensors, and retrieves these dimensions from all tokens' keys to determine the top-$k$ tokens for attention.
    \item \textbf{InfLLM}~\cite{DBLP:journals/corr/abs-2402-04617} organizes tensors into blocks and uses representative tokens to identify important blocks for attention.
\end{itemize}
Among the baselines, H2O, SnapKV, and PyramidKV are KVCache dropping methods, while SPARQ and InfLLM are KVCache offloading methods.
In addition, we consider an ``Oracle'' method that retrieves the exact top-$k$ tokens for each head. 
In our experiments, we align the number of tokens for selective attention and the data transfer amount in Oracle, SPARQ, InfLLM, and \name, to achieve a fair comparison.
In experiments on model quality, we allow KVCache dropping methods (H2O, SnapKV, PyramidKV) to attend to more tokens, matching the memory usage of the selected key-value pairs and the data transfer amount in the other methods, following the experiment settings in SPARQ.
We add a suffix of (C) to these methods, where ``C'' means compensation.

In our experiments, we specify two hyperparamters.
The first is the ratio of previous tokens that participate in selective attention.
For example, if the ratio is set to 1/5, we only use $s/5$ tokens for attention computation from an input sequence length of $s$.
The second hyper-paramter is the ratio of extra communication relative to the original tokens' keys, which we set to 1/128 in LongBench and 1/64 in InfiniteBench.
For \name, we set $m=2, b=6$ in LongBench, and set $m=4, b=8$ in InfiniteBench.
In each layer, we need to fetch PQ codes, which occupy $h_{kv}msb/8$ bytes, as each code only requires $b$ bits.
The original tokens' keys, using FP16, consume $2h_{kv}sd_h$ bytes.
In LongBench, since $(h_{kv}msb/8) / (2h_{kv}sd_h) = mb/(16d_h) = b/8 \times 1/128 \le 1/128$, \name satisfies the 1/128 ratio requirement.
In InfiniteBench, the calculation yields $mb/(16d_h)=1/64$.
For SPARQ, which retrieves $r$ values from $d_h=128$ values, we set $r=1$ for 1/128 and $r=2$ for 1/64.
For InfLLM, we use 1 or 2 representative token from every 128 tokens.

\subsubsection{\textbf{Hardware Environment and Hyperparameters}}

Unless otherwise specified, for each experiment, we use an NVIDIA GeForce RTX 4090 24GB card for GPU computation, two Intel(R) Xeon(R) Gold 6330 CPUs for K-Means clustering, 500GB CPU memory, and PCI-e 1.0 (x16) for communication.
For PQ construction in each layer, we have $m * h_{kv}$ clustering processes, each executed using 4 CPU threads.
For the GPU cache, we use 128 tokens per block to reduce management overhead.
For other hyperparameters, we align them with the settings from the corresponding papers or open-source codes.
It's important to note that in our experimental figures, ``k'' represents 1000, while ``K'' represents 1024.
In model performance experiments that require more memory than a single GPU card can provide, such as H2O (incompatible with FlashAttention), we utilize two or more GPU cards to partition the model.

\subsection{Model Performance}
\label{sec:expr:perf}

\subsubsection{\textbf{LongBench Evaluation}}
\label{sec:expr:perf:longbench}

The LongBench evaluation results on Llama-3.1-8B are presented in Table~\ref{tab:longbench:llama3}.
We experiment using 1/5 and 1/10 of the input tokens in selective attention, 
with an extra communication equivalent to 1/128 of the memory used by the previous tokens' keys.
Excluding Oracle, the best results for each setting, are highlighted in bold.
We also experiment on the Mistral-7B-inst-v0.2 model, with results presented in Appendix A.

\begin{table}[htbp]
    \scriptsize
    \setlength{\tabcolsep}{5pt}
    \caption{LongBench evaluation of the Llama-3.1-8B model (128K context length). 
    ``Ora'' stands for Oracle.
    ``H(C)'', ``S(C)'', and ``P(C)'' represent H2O, SnapKV, and PyramidKV, which are configured with extra tokens such that the memory usage matches that of other methods' selected tokens and transferred data amount.
    ``Inf'', ``SPA'', and ``PQC'' denote InfLLM, SPARQ, and \name, which all involve extra communications at an amount of 1/128 keys memory for pre-calculating relevance.}
    \label{tab:longbench:llama3}

    \newcommand{\cwidth}[0]{0.465cm}
    \begin{tabular}{|p{1.25cm}<{\centering}|p{\cwidth}<{\centering}|p{\cwidth}<{\centering}p{\cwidth}<{\centering}p{\cwidth}<{\centering}p{\cwidth}<{\centering}p{\cwidth}<{\centering}p{\cwidth}<{\centering}p{\cwidth}<{\centering}|p{\cwidth}<{\centering}p{\cwidth}<{\centering}p{\cwidth}<{\centering}p{\cwidth}<{\centering}p{\cwidth}<{\centering}p{\cwidth}<{\centering}p{\cwidth}<{\centering}|}
    \toprule
     &  &  \multicolumn{7}{c|}{\textbf{1/5 \#Tokens + 1/128 Extra Comm}} & \multicolumn{7}{c|}{\textbf{1/10 \#Tokens + 1/128 Extra Comm}} \\
    \multirow{-2}{*}{\textbf{Dataset}} & \multirow{-2}{*}{\textbf{Full}} & \textbf{Ora} & \textbf{H(C)} & \textbf{S(C)} & \textbf{P(C)} & \textbf{Inf} & \textbf{SPA} & \textbf{PQC} & \textbf{Ora} & \textbf{H(C)} & \textbf{S(C)} & \textbf{P(C)} & \textbf{Inf} & \textbf{SPA} & \textbf{PQC} \\
    \midrule
    NarrativeQA & 29.91 & 30.30 & \textbf{30.94} & 30.02 & 30.50 & 26.00 & 28.82 & 30.08 & 30.92 & 29.02 & 30.36 & \textbf{31.12} & 21.06 & 29.24 & 30.14 \\
    Qasper & 44.79 & 44.57 & 37.38 & 42.55 & 43.65 & 35.07 & 35.11 & \textbf{44.93} & 44.82 & 34.08 & 39.05 & 39.09 & 23.93 & 30.28 & \textbf{44.36} \\
    MultiFieldQA & 54.63 & 54.68 & 48.80 & 54.43 & 54.30 & 42.92 & 48.34 & \textbf{55.11} & 54.62 & 45.66 & 53.71 & 53.07 & 37.90 & 43.38 & \textbf{54.97} \\
    HotpotQA & 55.81 & 55.98 & 55.50 & \textbf{56.59} & 55.75 & 46.42 & 53.86 & 55.60 & 55.98 & 54.39 & 56.05 & 55.89 & 41.96 & 52.52 & \textbf{56.35} \\
    2WikiMQA & 45.78 & 46.16 & 45.23 & 46.02 & \textbf{46.32} & 27.81 & 45.24 & 45.89 & 46.59 & 44.23 & 45.04 & 45.61 & 23.66 & 45.40 & \textbf{45.79} \\
    Musique & 30.41 & 30.36 & 28.38 & 30.32 & 30.44 & 21.60 & 28.27 & \textbf{30.50} & 30.68 & 27.19 & \textbf{30.29} & 29.92 & 18.14 & 26.72 & 30.15 \\
    GovReport & 35.23 & 35.01 & 32.41 & 30.06 & 29.77 & 32.42 & 29.37 & \textbf{34.29} & 34.78 & 30.25 & 27.15 & 26.81 & 30.89 & 26.87 & \textbf{33.97} \\
    QMSum & 25.11 & 25.52 & 24.53 & 24.77 & 24.91 & 23.55 & 23.82 & \textbf{25.67} & 25.23 & 23.60 & 24.74 & 24.90 & 21.80 & 23.60 & \textbf{25.24} \\
    MultiNews & 27.30 & 26.70 & 26.28 & 24.26 & 23.40 & 24.98 & 24.42 & \textbf{26.33} & 26.78 & 25.30 & 22.65 & 22.08 & 23.48 & 22.78 & \textbf{25.95} \\
    TREC & 72.50 & 72.50 & 69.00 & 70.00 & 70.50 & 61.50 & 64.50 & \textbf{72.50} & 72.50 & 67.50 & 66.50 & 66.50 & 53.00 & 54.00 & \textbf{72.00} \\
    TriviaQA & 91.65 & 91.48 & 91.88 & 92.17 & \textbf{92.30} & 87.53 & 91.15 & 91.79 & 91.76 & 91.34 & 91.72 & \textbf{92.34} & 83.41 & 90.69 & 91.65 \\
    SAMSum & 43.80 & 43.57 & 42.68 & 42.89 & 42.71 & 42.99 & 43.27 & \textbf{43.66} & 43.95 & 42.73 & 42.38 & 41.94 & 42.77 & 42.76 & \textbf{43.34} \\
    Count & 6.72 & 7.22 & \textbf{6.81} & 6.72 & 6.72 & 3.00 & 6.22 & 6.72 & 7.22 & \textbf{6.72} & 6.72 & 6.72 & 4.00 & 6.23 & \textbf{6.72} \\
    Retrieval & 99.50 & 99.50 & 94.50 & \textbf{99.50} & \textbf{99.50} & 83.50 & 97.00 & 99.00 & 99.50 & 90.50 & \textbf{99.50} & \textbf{99.50} & 55.00 & 96.50 & \textbf{100} \\
    \midrule
    Average & 47.37 & 47.40 & 45.31 & 46.45 & 46.48 & 39.95 & 44.24 & \textbf{47.29} & 47.52 & 43.75 & 45.42 & 45.39 & 34.36 & 42.21 & \textbf{47.19} \\
    \bottomrule
    \end{tabular}
    
\end{table}

Compared to the baselines, \name achieves an improvement of +1.74\% using 1/5 \#tokens and +3.90\% using 1/10 \#tokens.
The offloading-based baseline methods, InfLLM and SPARQ, do not perform well, as we limit additional communication to minimize latency.
SnapKV(C) and PyramidKV(C), which assume that the last segment of the input context provides crucial token importance information, perform well in the benchmarks and tasks in this paper, particularly because the input questions are consistently positioned at the end of the context.
In later paragraphs, we conduct further experiments demonstrating that SnapKV(C) and PyramidKV(C) fail to match the performance of \name when questions are not positioned at the end.
Despite KVCache dropping baselines using more tokens for compensation and making specific assumptions on question positioning,
\name outperforms all these methods, demonstrating superior generation performance.
On most datasets, \name performs the best.
While a small number of tasks see some baseline methods slightly outperforming \name, the differences are not substantial.

The column ``Full'' shows model performance without compression.
However, it seems ``Oracle'', which uses ideally top-$k$ selective attention, performs even better.
There are mainly two reasons.
(1) In LongBench, when the sequence length exceeds the model's maximum context length, only the initial and the last tokens are used~\cite{DBLP:journals/corr/abs-2308-14508}.
(2) Selective attention only focuses on the tokens that really matters, which makes the generation perform better.
Compared to ``Oracle'', \name achieves comparable results, with an average difference of less than 0.70\%.
This difference is 6.36 times smaller than that of other baselines.
In some cases, we observe that \name even beats ``Oracle''. 
This suggests that clustering may help \name uncover intrinsic structures within the KVCache latent space, thus leading to promising results.
Furthermore, although it is usually expected that the performance should drop when there are fewer tokens, there are exceptions where the opposite happens. 
This could be because not all tokens are useful for generating new ones, so getting rid of unnecessary ones might enhance inference.

\begin{table}[tbhp]
    \scriptsize
    \caption{LongBench question-answering tasks evaluation using Llama-3.1-8B, with questions placed before other contexts. (1/10 \#Tokens, 1/128 extra communication)}
    \label{tab:longbench:variant}
    
    \newcommand{\cwidth}[0]{0.5cm}
    \begin{tabular}{|p{1.25cm}<{\centering}|p{\cwidth}<{\centering}|p{\cwidth}<{\centering}|p{\cwidth}<{\centering}|}
    \toprule
    \textbf{Dataset} & \textbf{S(C)} & \textbf{P(C)} & \textbf{PQC} \\
    \midrule
    NarrativeQA & \textbf{19.39} & 19.35 & 18.82 \\
    Qasper & 18.70 & 17.60 & \textbf{25.76} \\
    MultiFieldQA & 43.48 & 44.07 & \textbf{47.99} \\
    HotpotQA & 42.93 & 42.55 & \textbf{43.28} \\
    2WikiMQA & 36.36 & 35.04 & \textbf{36.48} \\
    Musique & 20.86 & 21.19 & \textbf{22.29} \\
    \midrule
    Average & 30.29 & 29.97 & \textbf{32.44} \\
    \bottomrule
    \end{tabular}
    
\end{table}

Since the strong baselines SnapKV(C) and PyramidKV(C) rely on the assumption that the last segment of the input, typically the question, can determine the important tokens, we experiment on an alternative setup where the questions are not placed at the end.
Specifically, we position the questions before the other contexts in the question-answering tasks of LongBench.
As shown in Table~\ref{tab:longbench:variant}, as we alter the positions of the input elements without further tuning the prompt, the overall results are inferior to those from the standard LongBench evaluation.
In this modified scenario, \name demonstrates superior performance over SnapKV(C) and PyramidKV(C) across nearly all datasets, achieving a significant improvement of +7.10\%.
Though SnapKV(C) and PyramidKV(C) perform well in the benchmarks and tasks evaluated in this paper where questions are typically positioned at the end of the input, their performance declines in scenarios with varied question placements — situations that frequently occur in real-world applications.
In contrast, \name, which does not rely on assumptions beyond selective attention, offers greater robustness and adaptability across various scenarios.

\subsubsection{\textbf{InfiniteBench Evaluation}}
\label{sec:expr:perf:infinitebench}

\begin{table}[tbhp]
    \scriptsize
    \setlength{\tabcolsep}{5pt}
    \caption{Infinitebench evaluation of the Llama-3.1-8B model (128K context length). 
    ``Ora'' stands for Oracle.
    ``H(C)'', ``S(C)'', and ``P(C)'' represent H2O, SnapKV, and PyramidKV, which are configured with extra tokens such that the memory usage matches that of other methods' selected tokens and transferred data amount.
    ``Inf'', ``SPA'', and ``PQC'' denote InfLLM, SPARQ, and \name, which all involve extra communications at an amount of 1/64 keys memory for pre-calculating relevance.}
    \label{tab:infinitebench}

    \newcommand{\cwidth}[0]{0.465cm}
    \begin{tabular}{|p{1.25cm}<{\centering}|p{\cwidth}<{\centering}|p{\cwidth}<{\centering}p{\cwidth}<{\centering}p{\cwidth}<{\centering}p{\cwidth}<{\centering}p{\cwidth}<{\centering}p{\cwidth}<{\centering}p{\cwidth}<{\centering}|p{\cwidth}<{\centering}p{\cwidth}<{\centering}p{\cwidth}<{\centering}p{\cwidth}<{\centering}p{\cwidth}<{\centering}p{\cwidth}<{\centering}p{\cwidth}<{\centering}|}
    \toprule
     &  &  \multicolumn{7}{c|}{\textbf{1/5 \#Tokens + 1/64 Extra Comm}} & \multicolumn{7}{c|}{\textbf{1/10 \#Tokens + 1/64 Extra Comm}} \\
    \multirow{-2}{*}{\textbf{Dataset}} & \multirow{-2}{*}{\textbf{Full}} & \textbf{Ora} & \textbf{H(C)} & \textbf{S(C)} & \textbf{P(C)} & \textbf{Inf} & \textbf{SPA} & \textbf{PQC} & \textbf{Ora} & \textbf{H(C)} & \textbf{S(C)} & \textbf{P(C)} & \textbf{Inf} & \textbf{SPA} & \textbf{PQC} \\
    \midrule
    En.Sum & 27.41 & 26.91 & 26.97 & 26.05 & 26.22 & 22.49 & 26.20 & \textbf{27.66} & 26.88 & 26.55 & 26.04 & 26.03 & 24.85 & 25.36 & \textbf{26.84} \\
    En.QA & 15.12 & 15.06 & 15.13 & 15.26 & 15.08 & 8.28 & 14.72 & \textbf{15.45} & 15.07 & 14.49 & 15.07 & 15.24 & 15.22 & \textbf{15.34} & 15.17 \\
    En.MC & 67.25 & 67.25 & \textbf{67.25} & \textbf{67.25} & \textbf{67.25} & 54.59 & \textbf{67.25} & \textbf{67.25} & 67.25 & 67.25 & 67.25 & 67.25 & 51.97 & \textbf{67.69} & 67.25 \\
    En.Dia & 16.50 & 16.50 & \textbf{21.00} & 16.50 & 14.00 & 9.00 & 15.50 & 17.00 & 18.50 & 19.00 & 16.50 & 13.00 & 12.00 & 15.50 & \textbf{21.50} \\
    Zh.QA & 13.05 & 13.10 & 13.04 & \textbf{13.49} & 12.94 & 10.56 & 13.08 & 13.05 & 13.28 & 12.56 & 13.31 & 13.15 & 11.29 & 12.48 & \textbf{13.41} \\
    Math.Find & 34.29 & 33.71 & 33.71 & \textbf{34.29} & \textbf{34.29} & 26.00 & \textbf{34.29} & 34.00 & 33.71 & 34.00 & \textbf{34.29} & \textbf{34.29} & 34.00 & \textbf{34.29} & 34.00 \\
    Retr.PassKey & 100 & 100 & \textbf{100} & \textbf{100} & \textbf{100} & 65.59 & \textbf{100} & \textbf{100} & 100 & 99.32 & \textbf{100} & \textbf{100} & \textbf{100} & \textbf{100} & \textbf{100} \\
    Retr.Number & 99.49 & 99.32 & 90.68 & \textbf{99.49} & \textbf{99.49} & 54.41 & 87.46 & 96.78 & 98.31 & 77.97 & 99.32 & 99.49 & \textbf{100} & 83.39 & 93.39 \\
    Retr.KV & 55.60 & 56.20 & 8.40 & 44.60 & 50.20 & 2.60 & 8.00 & \textbf{54.60} & 55.40 & 4.60 & 30.60 & 34.20 & 3.40 & 4.00 & \textbf{49.60} \\
    \midrule
    Average & 47.63 & 47.56 & 41.80 & 46.33 & 46.61 & 28.17 & 40.72 & \textbf{47.31} & 47.60 & 39.53 & 44.71 & 44.74 & 39.19 & 39.78 & \textbf{46.80} \\
    \bottomrule
    \end{tabular}
    
\end{table}

The InfiniteBench results on Llama-3.1-8B are presented in Table~\ref{tab:infinitebench}.
The experiment uses 1/5 and 1/10 of the input tokens in selective attention, with an extra communication equivalent to 1/64 of the memory used by the tokens' keys.
We use increased communication in InfiniteBench due to its much longer input contexts (around 100k).
Excluding Oracle, the best results for each setting, are highlighted in bold.

Compared to the baselines, \name performs the best and achieves an improvement of +1.50\% using 1/5 \#tokens and +4.60\% using 1/10 \#tokens.
Though SnapKV(C) and PyramidKV(C) perform well because questions are typically positioned at the end of the input context, \name still achieves superior performance.
On most datasets, \name performs the best and exhibits negligible degradation compared to Full and Oracle.
Although some baselines may perform better in specific tasks, \name consistently achieves scores that are very close. 
\name achieves results comparable to Oracle, with an average difference of less than 1.71\%.
This difference is 3.58 times smaller than that of other baselines.

\subsubsection{\textbf{Large-scale Needle-in-a-Haystack}}

\begin{figure}[tbhp]

\centering
\subfigure[Full.]{
\scalebox{0.23}{
\includegraphics[width=\linewidth]{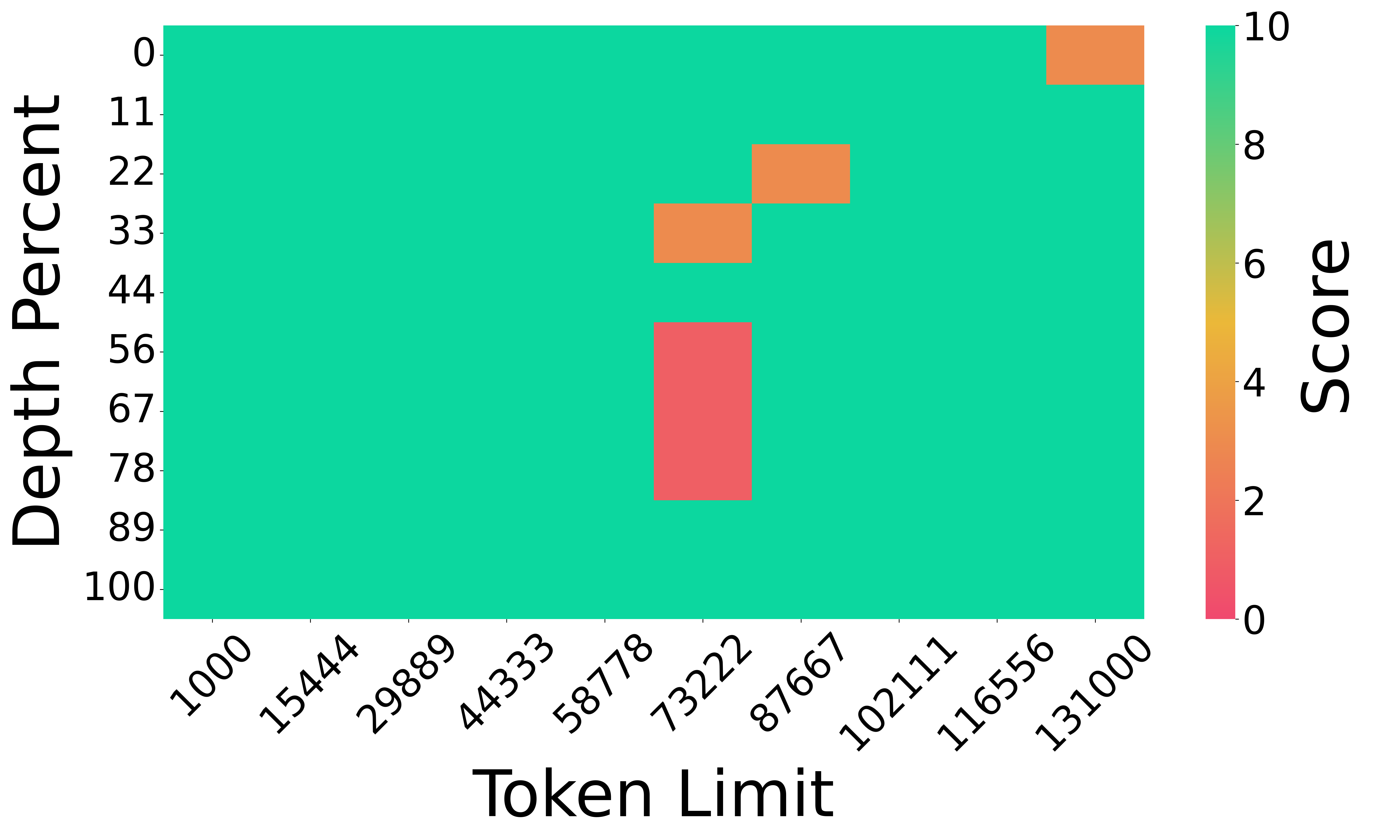}
}
}
\subfigure[Oracle.]{
\scalebox{0.23}{
\includegraphics[width=\linewidth]{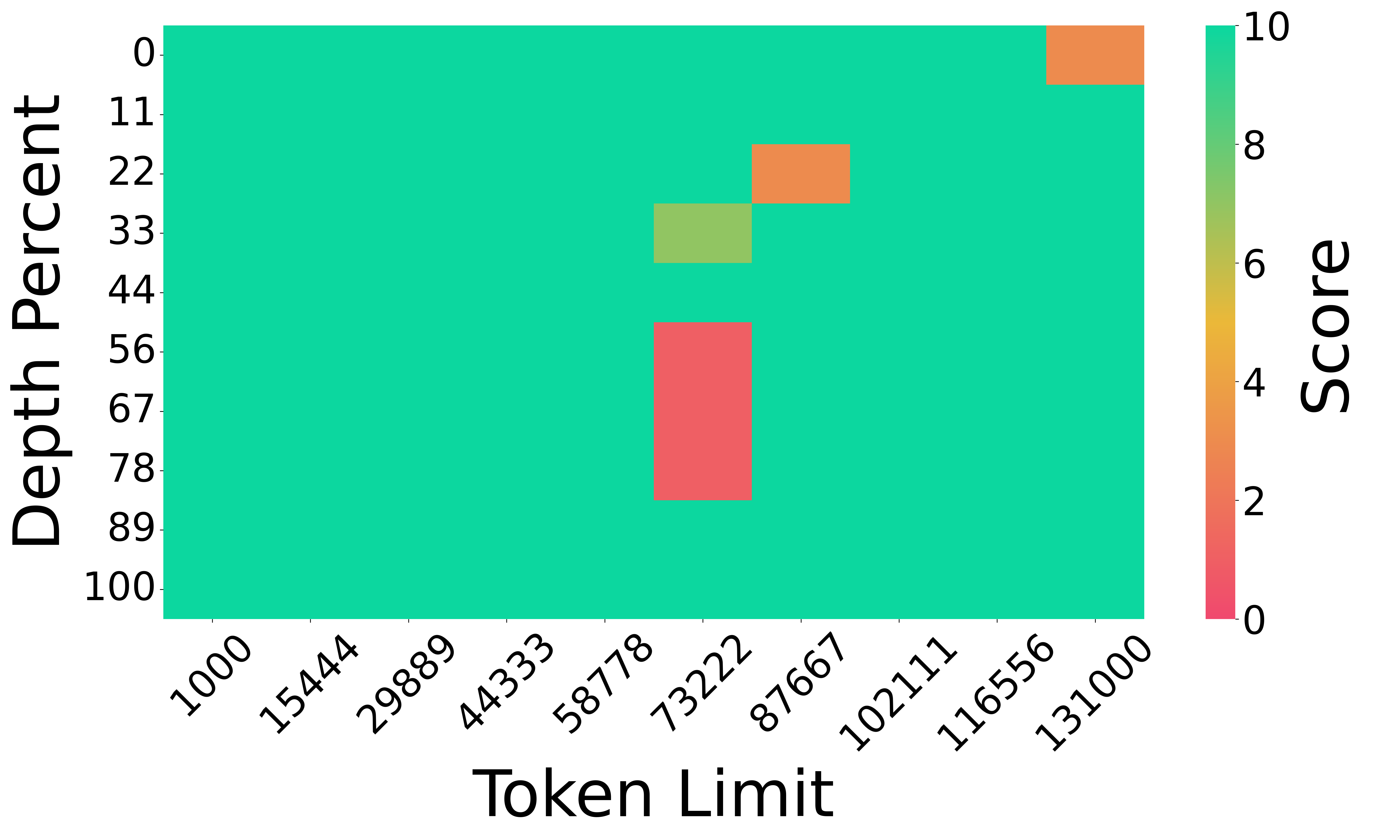}
}
}
\subfigure[H2O(C).]{
\scalebox{0.23}{
\includegraphics[width=\linewidth]{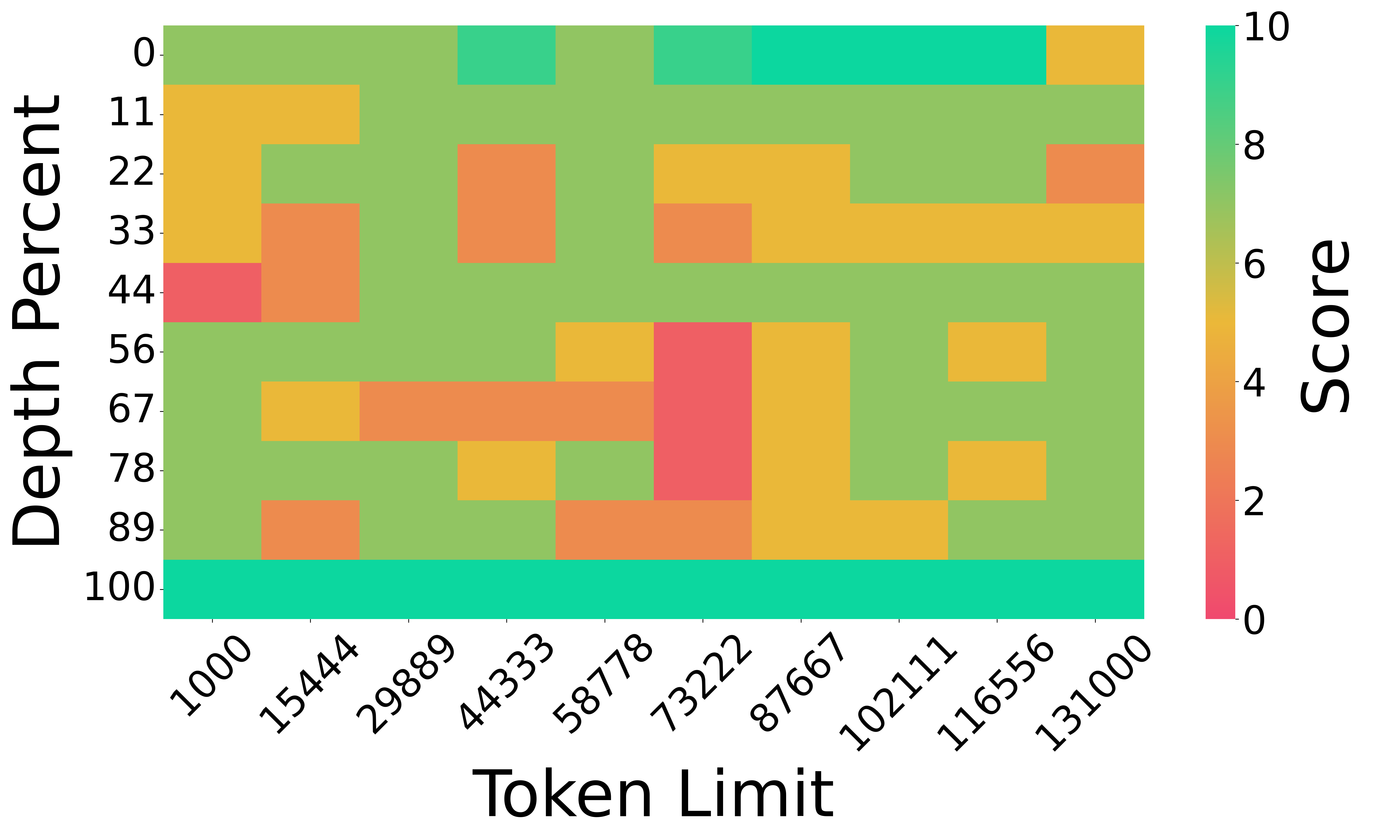}
}
}
\subfigure[SnapKV(C).]{
\scalebox{0.23}{
\includegraphics[width=\linewidth]{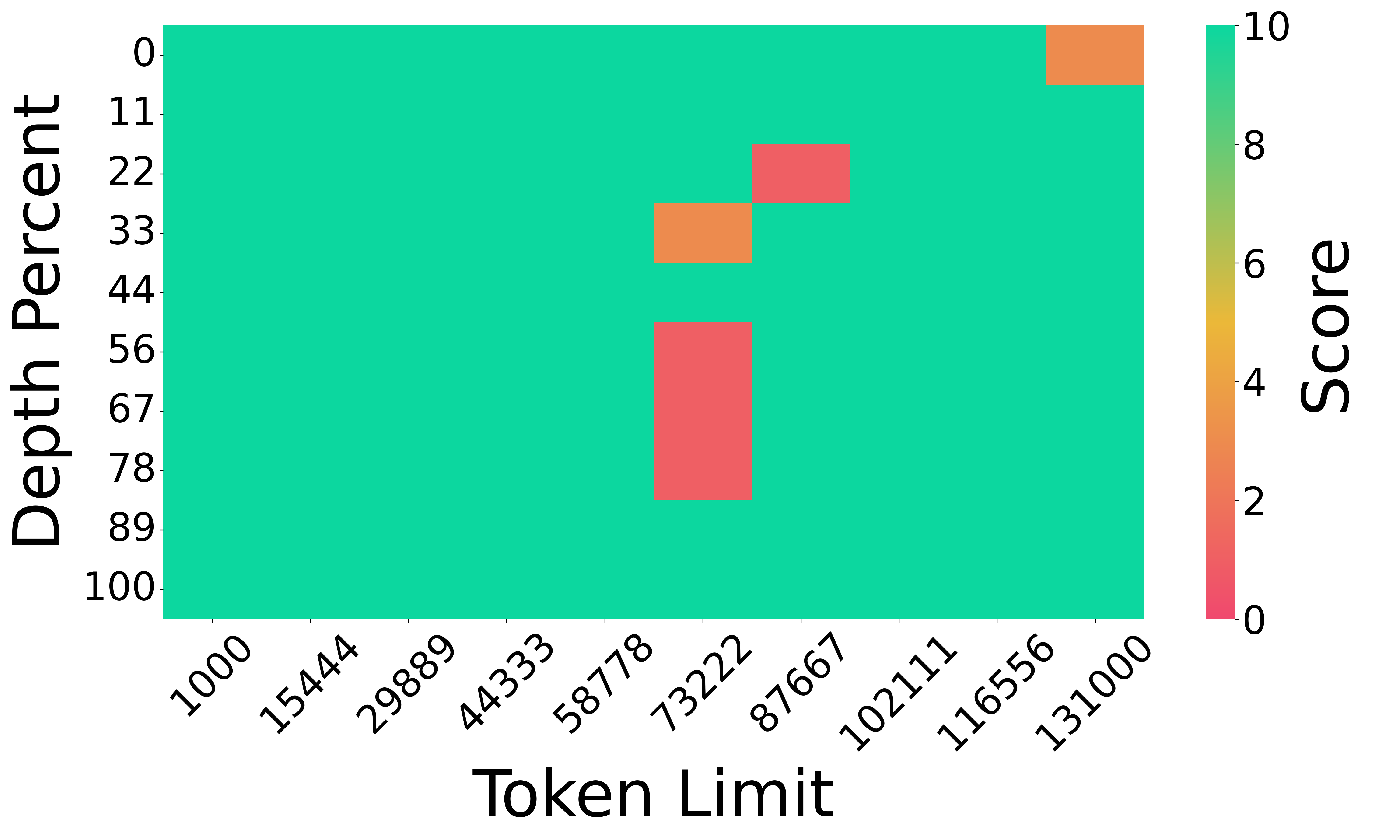}
}
}
\subfigure[PyramidKV(C).]{
\scalebox{0.23}{
\includegraphics[width=\linewidth]{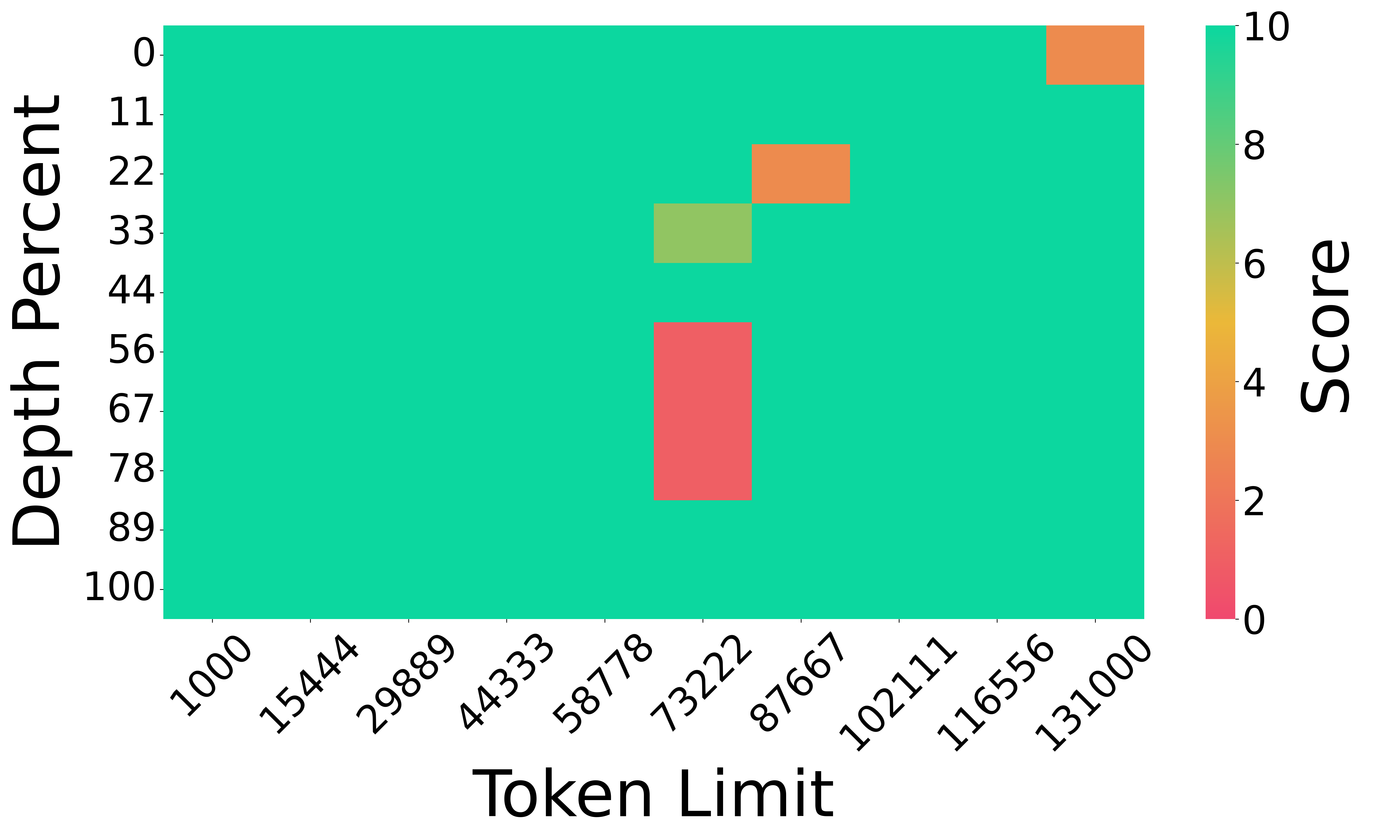}
}
}
\subfigure[SPARQ.]{
\scalebox{0.23}{
\includegraphics[width=\linewidth]{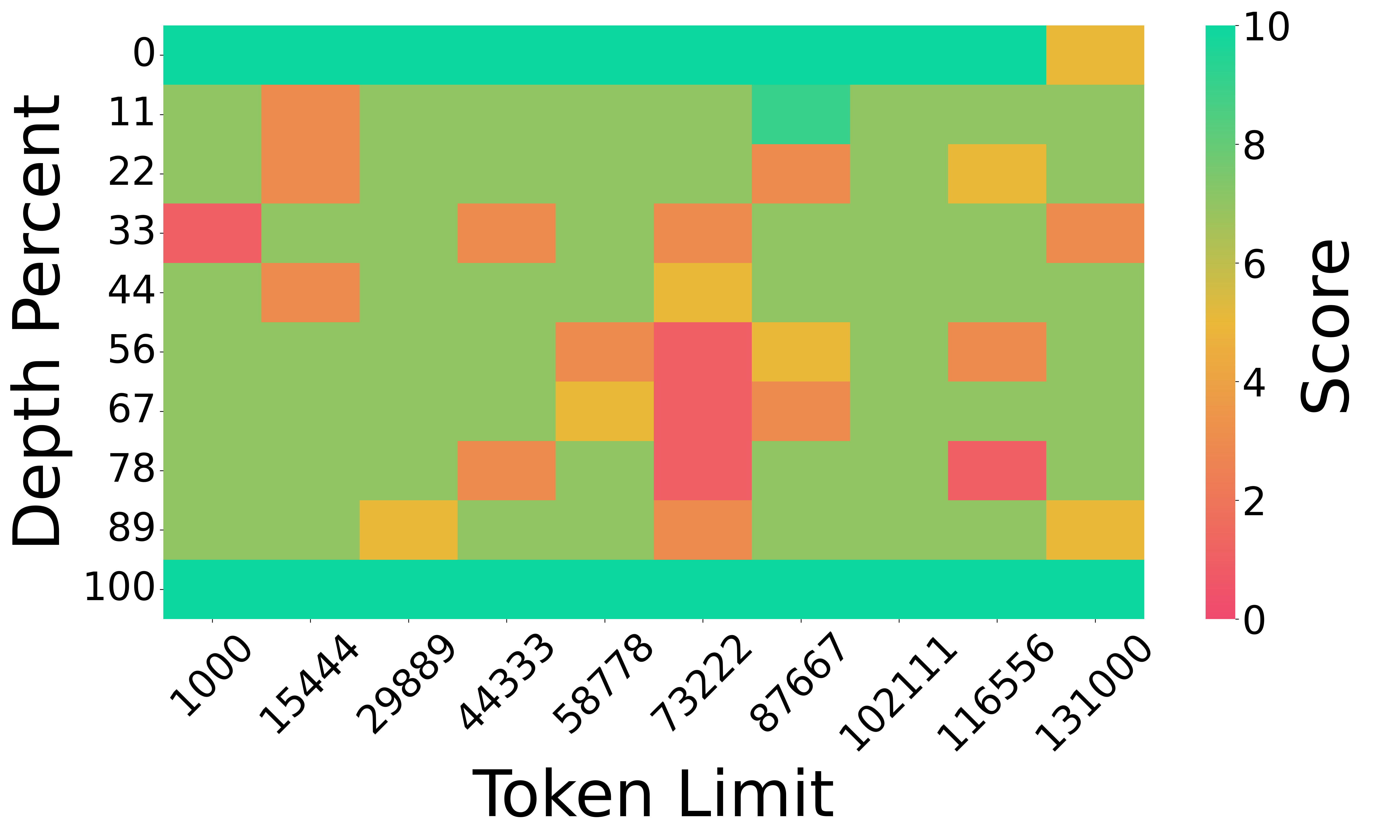}
}
}
\subfigure[InfLLM.]{
\scalebox{0.23}{
\includegraphics[width=\linewidth]{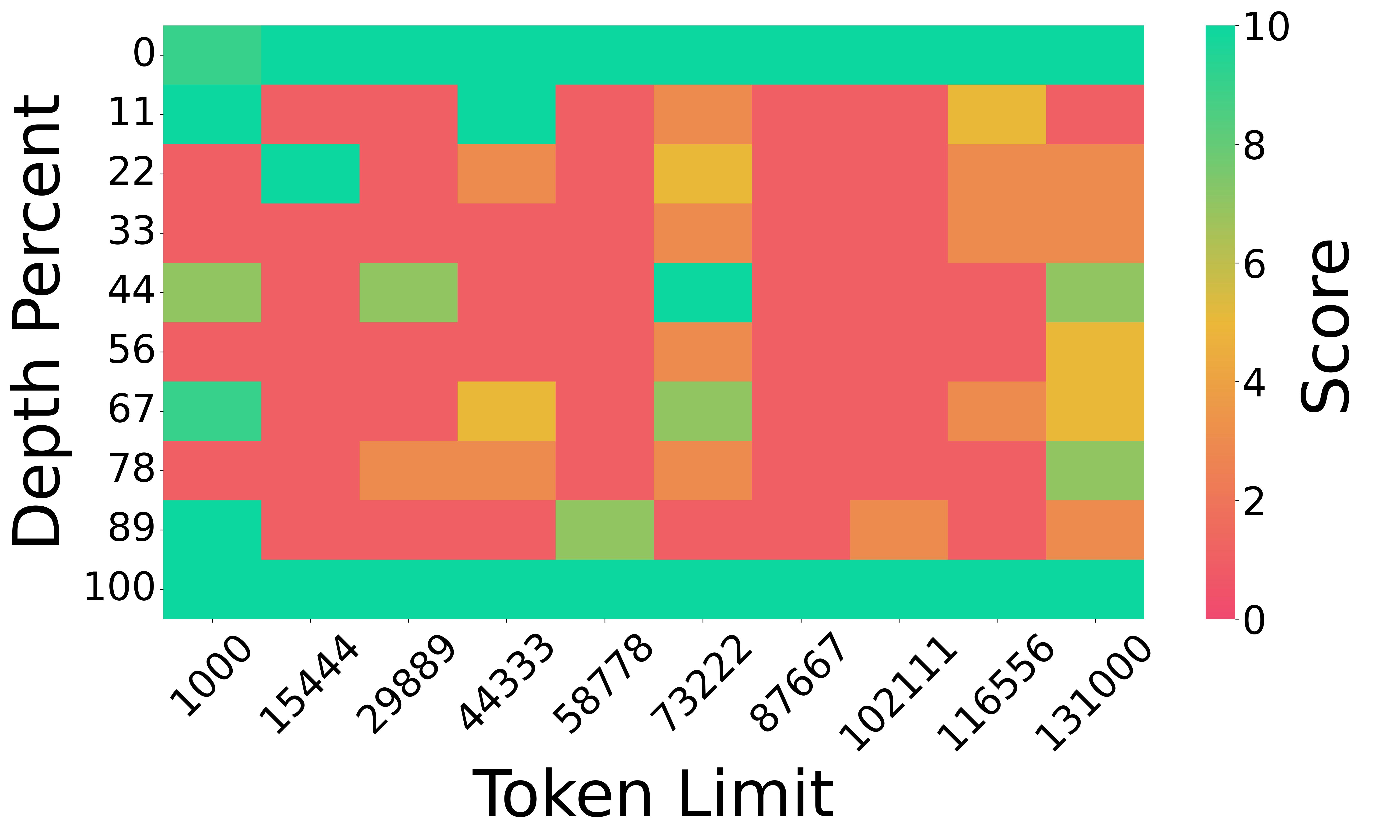}
}
}
\subfigure[\name.]{
\scalebox{0.23}{
\includegraphics[width=\linewidth]{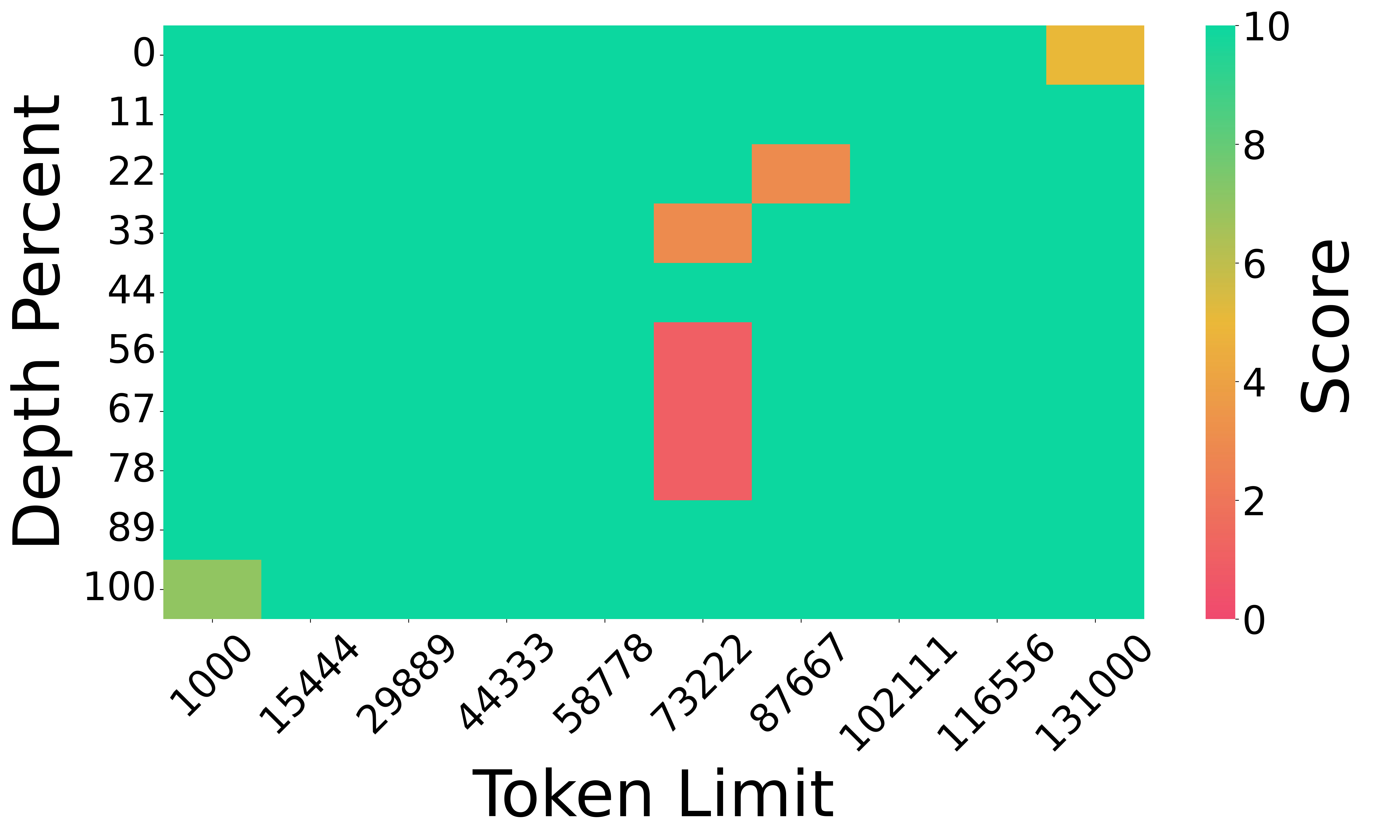}
}
}

\caption{Experimental results of the Needle-in-a-Haystack test.}
\label{fig:expr:perf:needle}

\end{figure}

We use the Llama-3.1-8B model to conduct the needle-in-a-haystack test.
We employ the common setting~\cite{longcontextprompting4claude}: the ``haystack'' is Paul Graham's Essays, and the ``needle'' is ``\textit{The best thing to do in San Francisco is eat a sandwich and sit in Dolores Park on a sunny day.}''
For each experiment, we use 1/10 the number of tokens in selective attention, and 1/64 extra communication. 
We use increased communication due to the long input contexts.
The results are shown in Figure~\ref{fig:expr:perf:needle}, where the $x$-axis represents the ``haystack'' length and the $y$-axis represents the position that the ``needle'' hides.
We significantly extend the maximum length to a large scale, specifically 131,000 tokens, as Llama-3.1-8B supports a maximum context length of $128 \times 1024 = 131,072$ tokens.
Greener shades indicate greater accuracy within the length range, demonstrating the model's ability to effectively retrieve the ``needle'' from the ``haystack''.

Among all the methods, SnapKV(C), PyramidKV(C), and \name demonstrate superior performance, successfully locating the needle in most of the scenarios.
These methods achieve performance nearly equivalent to Full and Oracle settings.
SnapKV(C) and PyramidKV(C) involve more tokens and take advantage of the evaluation setting where questions appear in the final segment of the input context.
\name, which involves minimal communication overhead and does not assume a specific position for the question, has shown promising results in large-scale scenarios.
Other baseline methods, however, fail to retrieve the needle in a substantial amount of cases.
InfLLM, in particular, struggles to find the needle in most cases, possibly due to its reliance on block-partitioning and the needle not being considered as representative tokens. 

\subsubsection{\textbf{Integrating with Prefilling Acceleration Method}}
\label{sec:expr:perf:minference}

Given that \name primarily enhances the decoding phase, we attempt to integrate it with MInference~\cite{DBLP:journals/corr/abs-2407-02490}, a state-of-the-art prefilling acceleration method. 
MInference designs sparse attention patterns during the prefilling phase to accelerate computation.
As we experiment the default settings of MInference, the results, as shown in Table~\ref{tab:minference}, indicate that MInference significantly degrades performance compared to dense attention baselines Full and \name.
After combining MInference and \name, we can simultaneously accelerate prefilling phase and address the KVCache memory issue in decoding phase.
Although faster prefilling shortens the clustering iterations of \name and sparse attention impacts the subsequent decoding process, \name still performs robustly. 
It exhibits only slight performance degradation compared to using MInference alone, demonstrating its strong compatibility with advanced prefilling acceleration methods.

\begin{table}[thbp]
    \scriptsize
    \caption{Performance of \name (1/5 \#tokens, 1/64 communication) combined with MInference on InfiniteBench using Llama-3.1-8B. ``PQC'' stands for PQCache, ``MInf'' stands for MInference, and ``Comb'' stands for their combination.}
    \label{tab:minference}

    \newcommand{\cwidth}[0]{0.7cm}
    \begin{tabular}{|c|p{\cwidth}<{\centering}|p{\cwidth}<{\centering}|p{\cwidth}<{\centering}|p{\cwidth}<{\centering}|}
    \toprule
    \textbf{Dataset} & \textbf{Full} & \textbf{PQC} & \textbf{MInf} & \textbf{Comb} \\
    \midrule
    En.Sum & 27.41 & 27.66 & 26.58 & 25.22 \\
    En.QA & 15.12 & 15.45 & 13.58 & 13.26 \\
    En.MC & 67.25 & 67.25 & 58.95 & 58.95 \\
    En.Dia & 16.50 & 17.00 & 12.50 & 11.50 \\
    Zh.QA & 13.05 & 13.05 & 11.85 & 11.93 \\
    Math.Find & 34.29 & 34.00 & 34.00 & 34.00 \\
    Retr.PassKey & 100 & 100 & 100 & 100 \\
    Retr.Number & 99.49 & 96.78 & 98.31 & 93.39 \\
    Retr.KV & 55.60 & 54.60 & 13.60 & 13.40 \\
    \midrule
    Average & 47.63 & 47.31 & 41.04 & 40.18 \\
    \bottomrule
    \end{tabular}
    
\end{table}

\subsubsection{\textbf{Results on Larger Model}}
\label{sec:expr:perf:largemodel}

To evaluate the performance of \name on larger LLMs, we conduct additional experiments using the Llama-3.1-70B model, with the results presented in Table~\ref{tab:largermodel}. 
We employ pipeline parallelism~\cite{DBLP:conf/nips/HuangCBFCCLNLWC19,guan2024advances} to partition the model layers on four 80G A800 GPUs.
We use 1/5 the number of tokens in selective attention and 1/128 extra communications.
For CPU resources, we simulate two settings: in one, we use the same amount of CPU resources per GPU as in the 8B model experiment; in the other, we use half the amount.
Still, the CPU computation employs limited clustering iterations to prevent GPU idle times and ensure efficiency.
Compared to the 8B model in Section~\ref{sec:expr:perf:longbench}, Llama-3.1-70B demonstrates superior performance, since it has much more parameters.
The performance gap between the non-compressed baseline and \name is negligible on the 70B model.
This can be attributed to two main factors:
(1) Larger model is equipped with enhanced capabilities and also demonstrates increased robustness to selective attention.
(2) When using a larger GQA model for \name, each layer experiences an increased computational burden on GPU, while the clustering task on CPU remains the same.
This is because in Llama, different model sizes have the same number of key-value heads\footnote{Note that this phenomenon generally applies to Llama-like models, although it may not be the case for some others, such as Qwen-2.5~\cite{qwen2.5}. We can employ the same analytical methodology for other models, and in most instances, the increase in PQ clustering computation is smaller than that of LLM computation as the model scales.}, which leads to the same complexity as analyzed in Section~\ref{sec:method:complexity}.
Therefore, by using the same amount of CPU resources per GPU and assuming that each GPU processes a layer simultaneously, \name is expected to have more clustering iterations and perform better with larger models.
As shown in Table~\ref{tab:largermodel}, even when we halve the CPU resources per GPU, \name already achieves performance comparable to the non-compressed baseline.

\begin{table}[tbhp]
    \scriptsize
    \caption{LongBench evaluation of Llama-3.1-70B (128K). We simulate half / same amount of CPU resources per GPU as in the experiments on 8B model.}
    \label{tab:largermodel}

    \newcommand{\cwidth}[0]{0.5cm}
    \begin{tabular}{|c|p{\cwidth}<{\centering}|p{\cwidth}<{\centering}p{\cwidth}<{\centering}|c|p{\cwidth}<{\centering}|p{\cwidth}<{\centering}p{\cwidth}<{\centering}|}
    \toprule
     &  &  \multicolumn{2}{c|}{\textbf{\name}} &  &  &  \multicolumn{2}{c|}{\textbf{\name}} \\
    \multirow{-2}{*}{\textbf{Dataset}} & \multirow{-2}{*}{\textbf{Full}} & Half & Same & \multirow{-2}{*}{\textbf{Dataset}} & \multirow{-2}{*}{\textbf{Full}} & Half & Same \\
    \midrule
    NarrativeQA & 35.07 & 35.02 & 35.02 & MultiNews & 26.95 & 26.77 & 27.16 \\
    Qasper & 49.97 & 49.25 & 49.02 & TREC & 76.50 & 76.50 & 76.00 \\
    MultifieldQA & 54.20 & 54.19 & 54.25 & TriviaQA & 94.04 & 94.04 & 94.04 \\
    HotpotQA & 64.95 & 65.07 & 64.95 & SAMSum & 47.37 & 47.67 & 47.47 \\
    2WikiMQA & 67.85 & 67.68 & 67.68 & Count & 20.00 & 20.00 & 20.00 \\
    Musique & 46.78 & 47.63 & 47.74 & Retrieval & 97.50 & 97.50 & 97.50 \\
    GovReport & 34.65 & 34.80 & 34.84 & &&& \\
    QMSum & 24.56 & 24.22 & 24.37 & \multirow{-2}{*}{\textbf{Average}} & \multirow{-2}{*}{\textbf{52.89}} & \multirow{-2}{*}{\textbf{52.88}} & \multirow{-2}{*}{\textbf{52.86}} \\
    \bottomrule
    \end{tabular}
\end{table}

\subsubsection{\textbf{GSM8k CoT Reasoning}}
\label{sec:expr:perf:cot}

We use the Mistral-7B-inst-v0.2 model for GSM8k CoT Reasoning.
Our prompt strategy involves 8 questions with 9-step reasoning and 2 questions with 8-step reasoning per sample, a common setup for long context inference~\cite{DBLP:journals/corr/abs-2305-17306}.
We use 1/128 extra communications.
As shown in Figure~\ref{fig:expr:perf:cot}, \name consistently outperforms H2O, SnapKV(C), PyramidKV(C), SPARQ, and InfLLM across different token counts, demonstrating strong reasoning capabilities.
Some results even surpass the uncompressed counterpart, suggesting that using part of the tokens can lead to improvements.
H2O(C) performs better than \name using 1/10 number of tokens, as it is allowed to access more tokens.

\subsubsection{\textbf{Impact of PQ Configuration}}
\label{sec:expr:perf:pqconf}

\begin{figure}[tbhp]
    \centering
    \subfigure[GSM8k CoT.]{
    \scalebox{0.23}{
    \includegraphics[width=\linewidth]{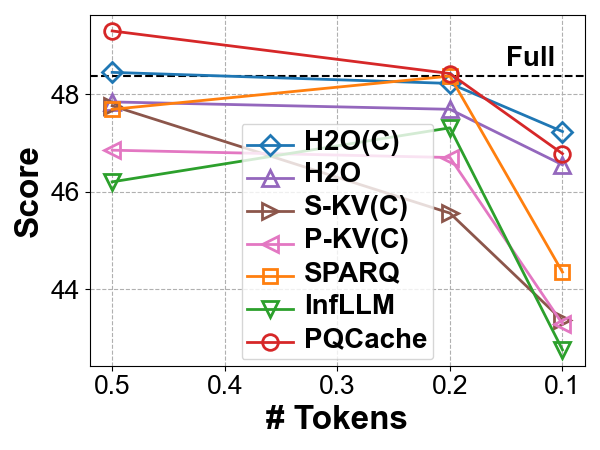}
    }\label{fig:expr:perf:cot}
    }
    \subfigure[PQ configurations.]{
    \scalebox{0.23}{
    \includegraphics[width=\linewidth]{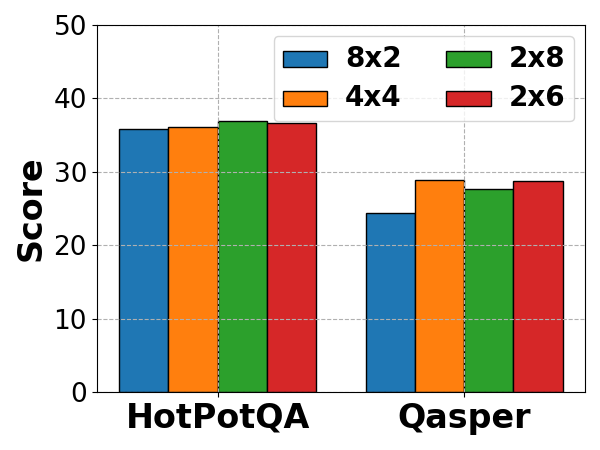}
    }\label{fig:expr:perf:pq}
    }
    \subfigure[Varying token number.]{
    \scalebox{0.23}{
    \includegraphics[width=\linewidth]{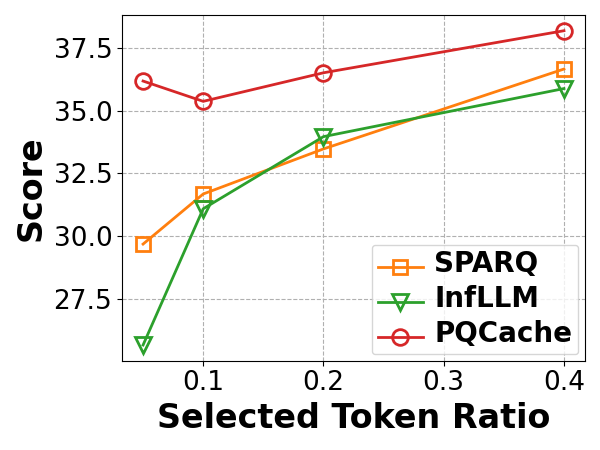}
    }\label{fig:expr:quality:numtoken}
    }
    \subfigure[Varying communications.]{
    \scalebox{0.23}{
    \includegraphics[width=\linewidth]{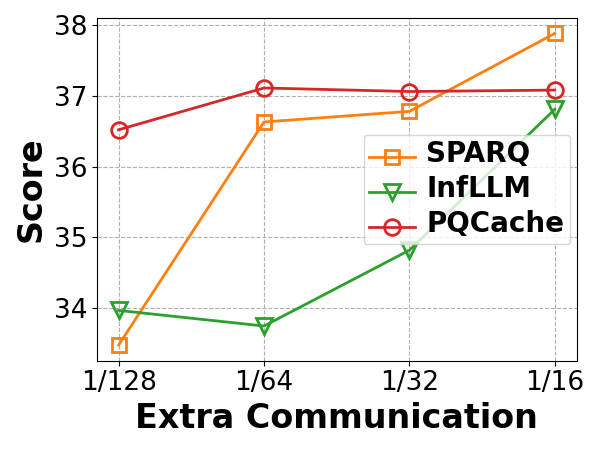}
    }\label{fig:expr:quality:extracomm}
    }
    \caption{(a) Performance on GSM8k with CoT. (b) Performance of \name with different PQ configurations. (c-d) Performance on HotPotQA under different settings.}
\end{figure}

We use Mistral-7B-inst-v0.2 to evaluate the effect of PQ configurations, including the number of partitions $m$ and the number of centroids $2^b$.
Figure~\ref{fig:expr:perf:pq} shows the results on HotPotQA and Qasper datasets with 1/10 tokens in selective attention, where the legends indicate $m\times b$.
The maximum number of clustering iterations is set as described in Section~\ref{sec:method:prefill} for each configuration, so as not to disturb GPU computation.
To ensure that expensive configurations do not benefit unfairly from the clipped maximum number of clustering iterations, we are experimenting with a more aggressive clipping strategy, which permits a very small number of clustering iterations.
As analyzed in Section~\ref{sec:expr:settings:baselines}, the involved communication is $mb/(16d_h)$.
Therefore, as long as $mb\le 16$, we can ensure that the extra communication will not exceed a size of 1/128 of tokens keys.
From the figure, we can see that \name is robust across various PQ configurations, with all configurations performing well except for the $8\times 2$ setting on Qasper dataset.
The configuration $2\times 6$ ($m=2$, $b=6$) yields the best performance.
It offers a sufficient number of centroids for representation, and in experiments it also has an appropriate number of clustering iterations.
Therefore, we choose it as the default configuration.

\subsubsection{\textbf{Variation in Token Number and Additional Communication}}

We use Mistral-7B-inst-v0.2 to investigate how the number of involved tokens and the amount of extra communication impact model performance on the HotPotQA dataset.
Fixing extra communication at 1/128 of the tokens' keys memory, we vary selected token ratio from 0.05 to 0.4.
As shown in Figure~\ref{fig:expr:quality:numtoken}, \name consistently outperforms the other baselines across different number of tokens.
All methods exhibit an upward trend as the number of tokens increases.
Fixing 1/5 of the tokens used, we vary the amount of extra communication from 1/128 to 1/16 of the tokens' keys memory.
As shown in Figure~\ref{fig:expr:quality:extracomm}, InfLLM and SPARQ has a upward trend when the communication increases, while \name remains stable.
Using 1/128 communication, the PQ structure is sufficient for \name to achieve a promising score.
In the 1/16 communication case, SPARQ achieves a slightly higher score than \name.
However, SPARQ already incurs significant latency under the 1/128 case, as shown in Section~\ref{sec:expr:efficiency:e2e}.
Even it performs well under the 1/16 case, the latency becomes increasingly unacceptable.
In low-communication scenarios, which are suitable for practical usage, \name achieves the best model performance.

\subsection{Efficiency}\label{sec:expr:efficiency}

In LLM inference, both prefilling and decoding efficiency are crucial.
When a user inputs a query, the LLM first performs prefilling to output the initial token, during which the user waits for a response. 
Therefore, lower prefilling latency is preferable.
During the decoding phase, tokens are generated one by one, and the speed should match human reading speed.

\subsubsection{\textbf{End-to-end Experiments}}
\label{sec:expr:efficiency:e2e}

\begin{figure}[tbhp]
    \centering
    \subfigure[Time to 2nd token.]{
    \scalebox{0.23}{
    \includegraphics[width=\linewidth]{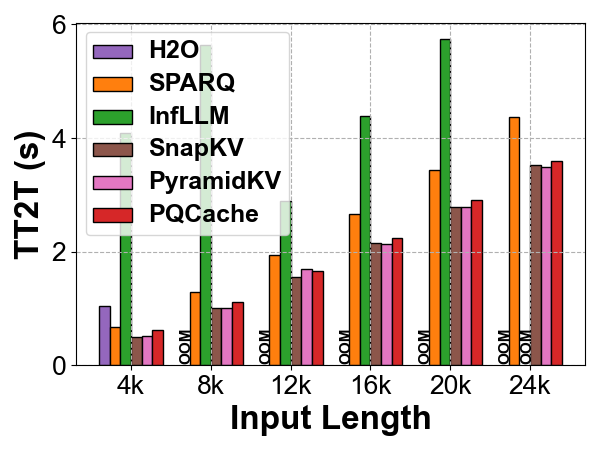}
    }\label{fig:expr:lat:tt2t}
    }
    \subfigure[Time per output token.]{
    \scalebox{0.23}{
    \includegraphics[width=\linewidth]{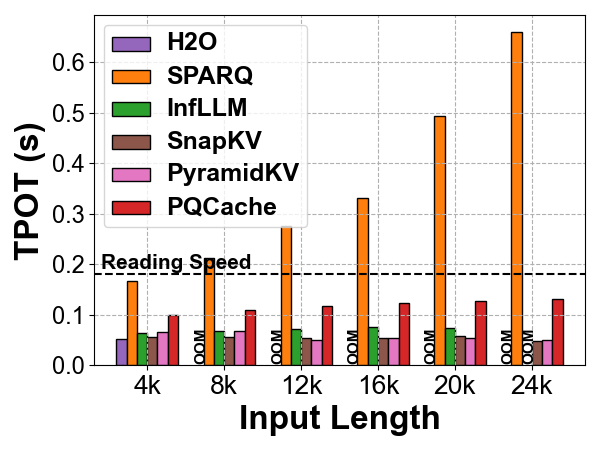}
    }\label{fig:expr:lat:tpot}
    }
    \subfigure[Cache improvements.]{
    \scalebox{0.23}{
    \includegraphics[width=\linewidth]{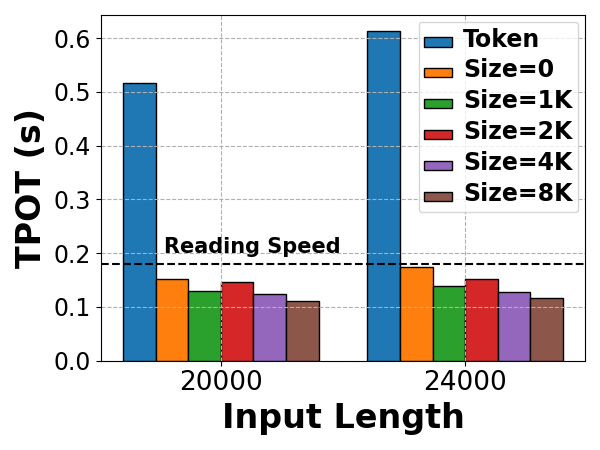}
    }\label{fig:expr:lat:cache:tpot}
    }
    \subfigure[Cache hit-rate.]{
    \scalebox{0.23}{
    \includegraphics[width=\linewidth]{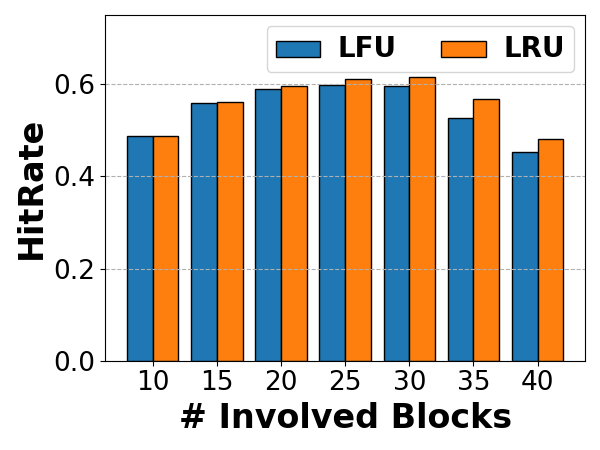}
    }\label{fig:expr:lat:cache:hitrate}
    }
    \caption{(a-b) Latency experiments. (c-d) GPU Cache experiments.}
    \label{fig:expr:lat}
\end{figure}

In \name, K-Means clustering runs concurrently with GPU computation, affecting the generation of the second token rather than the first.
Therefore, instead of evaluating Time To First Token (TTFT), we use Time To Second Token (TT2T) to assess system optimization.
This metric accounts for both the query entry to LLM output time and KVCache management overhead.
As shown in Figure~\ref{fig:expr:lat:tt2t}, with overlapping and adaptive clustering, \name achieves nearly the lowest TT2T compared to baseline methods.
H2O collects attention scores during prefilling, preventing the use of FlashAttention for acceleration and leading to OOM issues with lengthy input.
SnapKV and PyramidKV introduce negligible overhead during prefilling, resulting in low latency comparable to \name.
SPARQ, while having no prefilling overhead, suffers from a slow decoding process, resulting in higher TT2T.
InfLLM incurs time overhead due to the setup required for block-level KVCache management.

Time Per Output Token (TPOT) measures the time of each decoding step.
We present the TPOT of the methods in Figure~\ref{fig:expr:lat:tpot}.
Here we use 1/5 number of tokens with 1/128 extra communication in selective attention, and a GPU cache capable of holding 4096 tokens' keys and values.
To ensure a fair comparison, we allocate space in the GPU cache for our PQ centroids as well.
SPARQ must complete the query computations to determine which dimension to fetch from CPU, making the communication dependent on these computations.
The sequential processing leads to the highest latency among the methods, with the communication scaling linearly with the sequence length.
Except for SPARQ, all the other methods exhibit per-token latency faster than the human reading speed, which is around 250 words ($\approx$333 tokens) per minute~\cite{DBLP:conf/osdi/ZhongLCHZL0024}.
H2O, SnapKV, and PyramidKV avoid extra communications, while InfLLM and PQCache both leverage system optimizations to accelerate decoding.
InfLLM's block-level token management allows it to efficiently gather data from the CPU; however, this block-level assumption negatively impacts the model's overall quality.
\name incorporates prefetching and caching, achieving an acceptable TPOT while not degrading model quality.
It also maintains a nearly stable TPOT that does not scale with increasing sequence length.

\subsubsection{\textbf{Cache Performance}}
\label{sec:latency:cache}

Based on the experiment settings of the previous section, we conduct an ablation study to evaluate the effectiveness of the GPU cache and assess the cache hit-rate.
In Figure~\ref{fig:expr:lat:cache:tpot}, we present the Time Per Output Token (TPOT) using various cache sizes ranging from 0 to 8K, including an experiment with a token-level cache of 4K size.
Compared to the no-cache setting, the current block-level cache significantly reduces TPOT, decreasing time by 26.3\% and 32.8\% for 4K and 8K cache sizes, respectively.
We don't utilize token-level cache, since it incurs high management overhead.
We also plot the cache hit-rate for Least Recently Used (LRU) and Least Frequently Used (LFU) policies across different numbers of top-$k_{cache}$ blocks in Figure~\ref{fig:expr:lat:cache:hitrate}.
This experiment is conducted on the HotpotQA dataset, with 1/10 tokens in selective attention and 4K tokens in GPU cache.
Both LRU and LFU show similar performance, achieving around 0.5 hit-rate across different numbers of blocks.
As the block count increases, the hit-rate initially rises because more tokens are found within blocks.
However, it eventually declines as the block count exceeds the cache size, disrupting the cache update logic. 
In practice, we set the number of blocks to 32, matching the cache size and yielding a hit rate of approximately 0.6.

\subsubsection{\textbf{Time Decomposition}}

\begin{figure}[htbp]
    \centering
    \subfigure[Prefilling time.]{
    \scalebox{0.23}{
    \includegraphics[width=\linewidth]{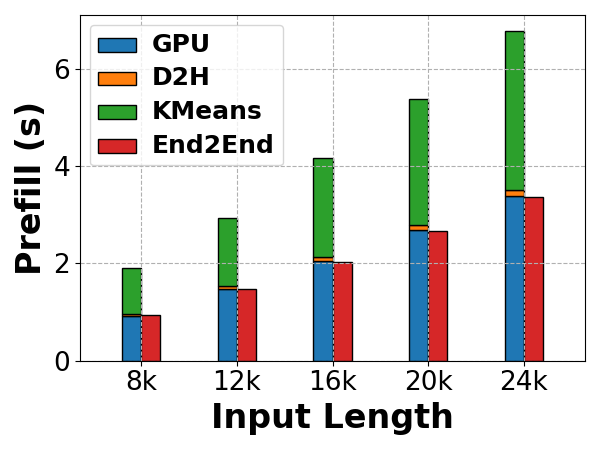}
    }\label{fig:expr:decomposition:prefill}
    }
    \subfigure[Decoding time.]{
    \scalebox{0.23}{
    \includegraphics[width=\linewidth]{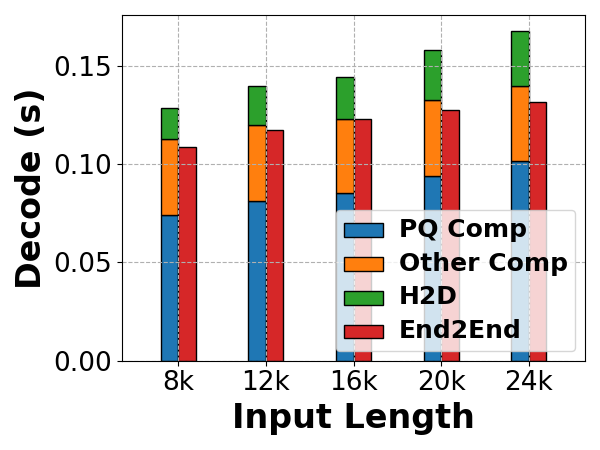}
    }\label{fig:expr:decomposition:decode}
    }
    \subfigure[Trade-off latency and score.]{
    \scalebox{0.46}{
    \includegraphics[width=0.5\linewidth]{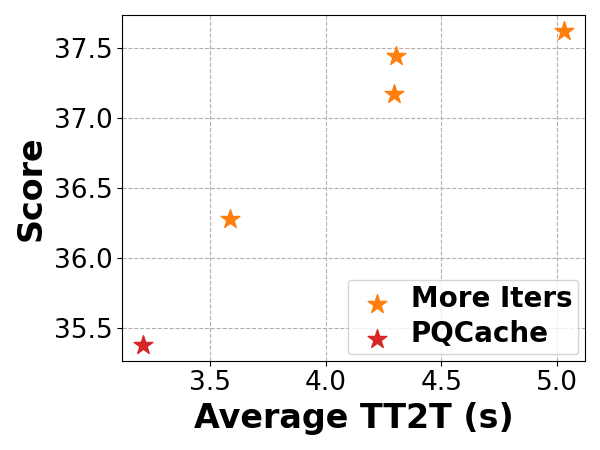}
    }\label{fig:expr:tradeoff:niter}
    }
    \caption{(a-b) Time decomposition experiments. (c) HotpotQA performance with varying KMeans iterations.}
    \label{fig:expr:decomposition}
\end{figure}

To verify our system improvements, we profile the time usage of LLM computation, PQ computation, and communication during both the prefilling and decoding phases.
Figure~\ref{fig:expr:decomposition:prefill} shows the profiled time of the prefilling phase.
The KVCache offloading time is negligible compared to GPU computation and KMeans time, as the prefilling stage is compute-intensive with $O(s^2)$ complexity.
With the adaptive number of clustering iterations in \name, the KMeans computation time closely matches the GPU computation time, demonstrating the accuracy and effectiveness of our cost model.
The end-to-end time for the entire prefilling phase, including GPU computation and KMeans, is also close to the GPU computation time due to perfect overlap of GPU computation, offloading, and KMeans processes.

Figure~\ref{fig:expr:decomposition:decode} shows the profiled time of the decoding phase.
We decompose the decoding phase by profiling the PQ computation time (including top-$k$ token index time), LLM computation time, and CPU-to-GPU communication time (including PQ codes and top-k tokens' key-value pairs) without our GPU cache optimization.
We also profile the end-to-end time with all optimizations. 
By using overlapping and GPU cache for acceleration, the PQ communication can be overlapped, and the token index time and top-$k$ tokens communication time can be reduced, resulting in an end-to-end time smaller than the sum of all components.
Our decoding time remains stable with increasing input length.

\subsubsection{\textbf{Number of Iterations in K-Means Clustering}}
\label{sec:expr:kmeans}

In \name, we design an adaptive K-Means clustering strategy to avoid blocking GPU computation.
To investigate its impact on model accuracy, we conduct experiments with varying numbers of clustering iterations on the HotpotQA dataset, using Mistral-7B-inst-v0.2 and 1/10 of the tokens in attention.
As shown in Figure~\ref{fig:expr:tradeoff:niter}, more clustering iterations generally improve accuracy but increase latency.
Though the adaptive strategy has a relatively low score, it achieves the least Time To 2nd Token (TT2T) by not blocking GPU computation, and already performs well.
Our experiments reveal the significant potential of integrating PQ into LLM inference, achieving superior performance when clustering is unrestricted. 
Additionally, \name offers flexibility to balance model quality and efficiency, providing practical options for various applications. 
For instance, in scenarios like mobile search, users may tolerate 7-10 seconds of latency\footnote{\textit{Impact of Response Latency on User Behaviour in Mobile Web Search}, Falk Scholer et al., CHIIR'21}, but model performance is always crucial. 
We provide an interface that allows users to set the number of iterations, enabling them to balance model performance and latency according to their specific needs.

\section{Discussion}
\label{sec:discussion}

\subsubsection*{\textbf{Hyper-parameter Tuning}}

In \name, several hyperparameters require tuning.
For the retrieval set size (top-$k$),  intuitive and empirical evidence from previous subsections suggests that larger set sizes typically yield better performance.
However, the practical limit is the available GPU memory.
Practitioners should adjust this parameter based on their specific workloads and hardware capabilities.
In the PQ settings, the number of iterations is automatically determined as detailed in Section~\ref{sec:method:prefill}.
Tuning the number of clusters and segments is common and necessary in all PQ-based methods in both ANNS~\cite{DBLP:journals/pami/JegouDS11,DBLP:journals/tbd/JohnsonDJ21} and deep learning scenarios~\cite{DBLP:conf/icml/ChenLS20,DBLP:conf/sigir/XiaoLHZLGCYSSX22}.
Here we provide a simple guideline for tuning.
We suggest starting with the product of $m$ and $b$, as it represents the size of the total vector space $(2^b)^m = 2^{mb}$.
The optimal scale should be neither too small to compromise vector representation nor too large to overburden the clustering process. 
Once an optimal $m\times b$ is found, minor adjustments can be made to refine the settings.
As demonstrated in Section~\ref{sec:expr:perf:pqconf}, performance varies slightly when maintaining a constant $m\times b$.

\subsubsection*{\textbf{Multiple Requests and Multiple GPUs}}

\name can be directly applied to serve multiple requests.
All computations and communications involved in \name, as detailed in Section~\ref{sec:method}, can be extended with a batch size dimension.
The linear increase can be managed through parallel computation, thereby maximizing GPU utilization when a single request cannot fully utilize it.
However, if the CPU is already compute-intensive, increasing the batch size might cause the CPU to block GPU computation.
A simple solution is to add more CPUs, given their much lower cost compared to current GPUs.
This solution is also applicable for multiple GPUs scenarios, where GPU computational capability is further enhanced.

\subsubsection*{\textbf{Longer Output Sequences and Multiple-turn Conversation}}

At present, \name clusters once for PQ construction, which is suitable for most LLM inference scenarios with relatively short output sequences.
We also assign new PQ codes to newly generated tokens based on their nearest centroids.
However, in scenarios with longer outputs, input-based PQ structures may not capture new information from output tokens.
A straightforward solution is to periodically reconstruct PQ to update the information.
LLMs are also commonly used in multiple-turn conversations.
In such cases, two strategies can be applied: (1) perform a prefill again upon each user input with all previous tokens; (2) perform K-Means on each input separately and add the output's PQ codes to the corresponding input.
The choice of strategy can be tailored to the request, depending on whether the multiple turns are related.

\subsubsection*{\textbf{GPU and CPU Computational Capability}}

In \name, model performance can sometimes be constrained by CPU computational capability, as we limit the number of clustering iterations on the CPU to prevent blocking GPU computation. 
In practice, LLM inference is often assigned to less computationally capable GPU cards, reserving more powerful cards for compute-intensive training. 
If the power of the GPU decreases, the K-Means clustering can have more iterations, which could potentially enhance the performance of \name in PQ construction.
Therefore, \name is highly compatible with LLM inference.
PQCache is ideally suited for environments with a proper GPU-to-CPU ratio, enabling effective PQ clustering to seamlessly overlap with LLM computation, regardless of whether the deployment is on the edge or in the cloud.

\subsubsection*{\textbf{Combined with LLM Inference/Serving Systems}}

Several LLM inference/serving systems, including Orca~\cite{DBLP:conf/osdi/YuJKKC22}, vLLM~\cite{DBLP:conf/sosp/KwonLZ0ZY0ZS23}, and DistServe~\cite{DBLP:conf/osdi/ZhongLCHZL0024}, have been proposed to enhance the overall efficiency.
\name, which aims to use fewer tokens for attention, is orthogonal and compatible with these systems.
As discussed in previous subsections, \name is compatible with batching techniques such as Orca~\cite{DBLP:conf/osdi/YuJKKC22}.
vLLM~\cite{DBLP:conf/sosp/KwonLZ0ZY0ZS23}'s block-level PagedAttention can be seamlessly integrated into \name, also compatible with our block-level GPU cache.
For systems like DistServe~\cite{DBLP:conf/osdi/ZhongLCHZL0024} that use prefill-decode disaggregation, \name only introduces new PQ structures communication, which is much smaller than the KVCache.
While combining \name with these systems may require some engineering efforts, we anticipate the practical application of \name in contemporary LLM inference systems.

\subsubsection*{\textbf{Retrieval for LLM Inference}}

To the best of our knowledge, our work is the first to incorporate information retrieval techniques into LLM inference.
In LLM inference, especially in long-context scenarios, the generation of the next token naturally depends on relevant contexts, making the integration of retrieval techniques both intuitive and effective.
We are proud to pioneer a new domain for efficient LLM inference and foresee the incorporation of retrieval as a standard paradigm in next-generation LLM inference. 
In \name, we currently utilize classic PQ structures for efficient LLM.
Other retrieval techniques, such as IVF~\cite{DBLP:journals/tbd/JohnsonDJ21} and HNSW~\cite{DBLP:journals/pami/MalkovY20}, along with recent advancements, could potentially contribute to more efficient LLM inference. 
We look forward to seeing more research exploring this new domain, leveraging and designing effective methods for improved LLM inference.

\section{Related Work}

\subsubsection*{\textbf{Selective Attention for KVCache}}
To eliminate the impact of memory-intensive KVCache, a group of methods include only essential tokens for attention computation during LLM inference.
One way is to discard unnecessary tokens.
LM-Infinite~\cite{DBLP:journals/corr/abs-2308-16137} and Streaming-LLM~\cite{DBLP:journals/corr/abs-2309-17453} only preserve the initial tokens and the most recent tokens.
H2O~\cite{DBLP:conf/nips/Zhang00CZC0TRBW23} and Scissorhands~\cite{DBLP:conf/nips/LiuDLWXXKS23} utilize attention scores to identify important tokens.
Their following works~\cite{DBLP:journals/corr/abs-2310-01801,DBLP:journals/corr/abs-2403-09054,wang2024squeezeattention,DBLP:journals/corr/abs-2402-06262} have explored adaptive token selection and additional metrics for better model accuracy.
The LLMLingua series~\cite{DBLP:conf/emnlp/JiangWLYQ23,DBLP:journals/corr/abs-2403-12968} leverage an auxiliary small model to tell which tokens are necessary. 
Since token-level compression evicts the tokens in a greedy manner, the information loss in subsequent decoding phase may lead to model degradation.
Another way is to fetch relevant tokens on demand during the decoding phase.
SPARQ~\cite{DBLP:journals/corr/abs-2312-04985} and InfLLM~\cite{DBLP:journals/corr/abs-2402-04617} offload KVCache to CPU, and selectively fetch relevant key-value pairs for each attention computation.
\name also falls under this category of methods, demonstrating effective and efficient LLM inference in comparison to existing techniques.

\subsubsection*{\textbf{KVCache Quantization}}
Quantization can be directly applied on the entire KVCache~\cite{DBLP:journals/corr/abs-2402-02750,DBLP:journals/corr/abs-2403-04643,DBLP:journals/corr/abs-2401-18079} - a straight-forward approach with promising model quality.
Other compression techniques can also be employed to address the residuals introduced by quantization~\cite{kang2024gear}.
Quantization is orthogonal to token importance, and recent research has explored applying both techniques~\cite{DBLP:journals/corr/abs-2402-18096}.

\subsubsection*{\textbf{KVCache Scheduling}}
Another related research direction is to meticulously schedule the KVCache within memory hierarchy.
FlexGen~\cite{DBLP:conf/icml/0007ZYLRCLRSZ23} employs linear programming to schedule the communication, searching for efficient patterns to store and access tensors.
AttentionScore~\cite{DBLP:journals/corr/abs-2403-19708} maintains a hierarchical KV caching system, allowing efficient reuse of KVCache across multi-turn conversations.
KVCache streaming for LLM serving~\cite{DBLP:journals/corr/abs-2310-07240,strati2024d} involves handling multiple requests within more levels of memory hierarchy.

\subsubsection*{\textbf{Embedding Management}}
Embedding vector management is a common research focus within the database and data management domains, including embedding compression~\cite{DBLP:journals/pvldb/ZhangZMSLYC23,zhang2024cafe,DBLP:conf/kdd/ShiMNY20}, embedding retrieval~\cite{DBLP:journals/pvldb/WangXY021,DBLP:journals/pvldb/HuangSLPKY20}, and key-value storage~\cite{DBLP:conf/damon/ChenZXQLLZ21,DBLP:journals/pvldb/RenZAG17}.
Our work provides a potential direction for integrating classic embedding management methods into the LLM ecology. 

\section{Conclusion}

In this paper, we proposed \name, a system-algorithm co-designed method for effective and efficient long context LLM inference.
We incorporated the embedding retrieval technique PQ to reduce both memory and computation burden, and leveraged PQ codes and centroids to facilitate efficient ANN search for important tokens used in the attention module.
Through meticulous overlapping and caching, we managed to minimize overhead to a negligible level.
We evaluated \name through extensive experiments and showed that it improved model quality by 4.60\% on InfiniteBench compared to existing methods, while maintaining low system latency.

\begin{acks}
This work is supported by National Science and Technology Major Project (2022ZD0116315), National Natural Science Foundation of China (U23B2048, U22B2037, 62402011), Beijing Municipal Science and Technology Project (Z231100010323002), China National Postdoctoral Program for Innovative Talents (BX20230012), China Postdoctoral Science Foundation (2024M750103), Beijing Natural Science Foundation (4244080), research grant No. IPT-2024JK29, and High-performance Computing Platform of Peking University.
\end{acks}

\bibliographystyle{ACM-Reference-Format}
\bibliography{reference}

\end{document}